\documentclass{article}

\usepackage[final]{neurips_2025}

\usepackage[utf8]{inputenc} % allow utf-8 input
\usepackage[T1]{fontenc}    % use 8-bit T1 fonts
\usepackage{hyperref}       % hyperlinks
\usepackage{url}            % simple URL typesetting
\usepackage{booktabs}       % professional-quality tables
\usepackage{amsfonts}       % blackboard math symbols
\usepackage{nicefrac}       % compact symbols for 1/2, etc.
\usepackage{microtype}      % microtypography
\usepackage{amsmath,amsfonts,amssymb}
\usepackage{algorithm}
\usepackage{algpseudocode}
\usepackage{array}

\usepackage{textcomp}
\usepackage{stfloats}
\usepackage{url}
\usepackage{soul}
\usepackage{verbatim}
\usepackage{graphicx}
\usepackage{multirow}
\usepackage{pifont}
\usepackage{tikz}
\usepackage{comment}
\usepackage{amsmath,amssymb} % define this before the line numbering.
\usepackage{colortbl}  %彩色表格需要加载的宏包
\usepackage{color}
\usepackage{multirow}
\usepackage{makecell}
\usepackage{booktabs}       % professional-quality tables
\usepackage{bm}

\usepackage{graphicx}
\usepackage{caption}
\usepackage{wrapfig}

\usepackage{caption}
\usepackage{stackengine}

\definecolor{myPurple}{RGB}{128,0,128}
\definecolor{myBlue}{RGB}{100, 150, 255}
\usepackage[capitalize]{cleveref}
\usepackage{caption}
\usepackage{colortbl}
\usepackage{booktabs}
\usepackage{marvosym}
\usepackage{lipsum}
\usepackage{xspace}
\usepackage{footmisc}
\usepackage{subcaption} % 加在导言区
\usepackage{xcolor}

\usepackage{colortbl}
\usepackage{array}

\definecolor{color3}{rgb}{0.95,0.95,0.95}
\definecolor{color4}{rgb}{0.90,0.9,0.9}

\title{UltraHR-100K: Enhancing UHR Image Synthesis with A Large-Scale High-Quality Dataset}
\author{Chen Zhao, En Ci, Yunzhe Xu, Tiehan Fan, Shanyan Guan, Yanhao Ge, Jian Yang, and Ying Tai
}
\author{%
	Chen Zhao\textsuperscript{1,}\thanks{Equal Contribution.},
	En Ci\textsuperscript{1,*},
	Yunzhe Xu\textsuperscript{1,*},
	Tiehan Fan\textsuperscript{1},
	Shanyan Guan\textsuperscript{2}, \\
	\And
	Yanhao Ge\textsuperscript{2},
	Jian Yang\textsuperscript{1},
	Ying Tai\textsuperscript{1,}\thanks{Correspondence to: Ying Tai.} \\
	\AND \\
	\centering  % 让后续内容居中
	\begin{tabular}{l}
		\textsuperscript{1} State Key Laboratory of Novel Software Technology, Nanjing University, China \\
		\textsuperscript{2} vivo Mobile Communication Co., Ltd., China \\
	\end{tabular}
}

\begin{document}

\maketitle

        \vspace{-6mm}
\begin{figure}[h!]
    \centering
    \begin{minipage}[t]{0.58\linewidth}
        \centering
        
        \includegraphics[width=\linewidth]{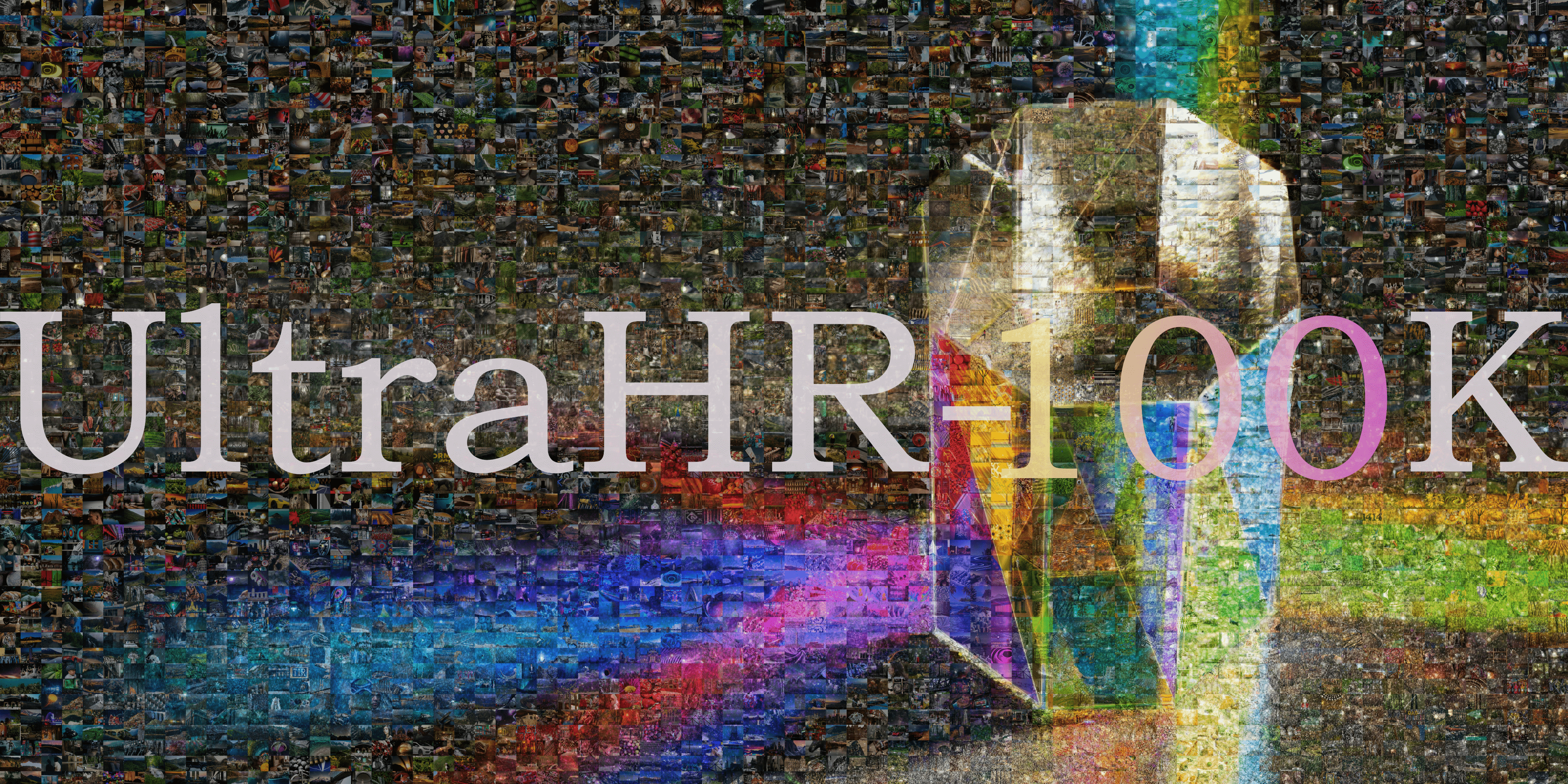}
        \vspace{-2mm}
    \end{minipage}%
    \hfill
    \begin{minipage}[t]{0.39\linewidth}
        \centering
        \includegraphics[width=\linewidth]{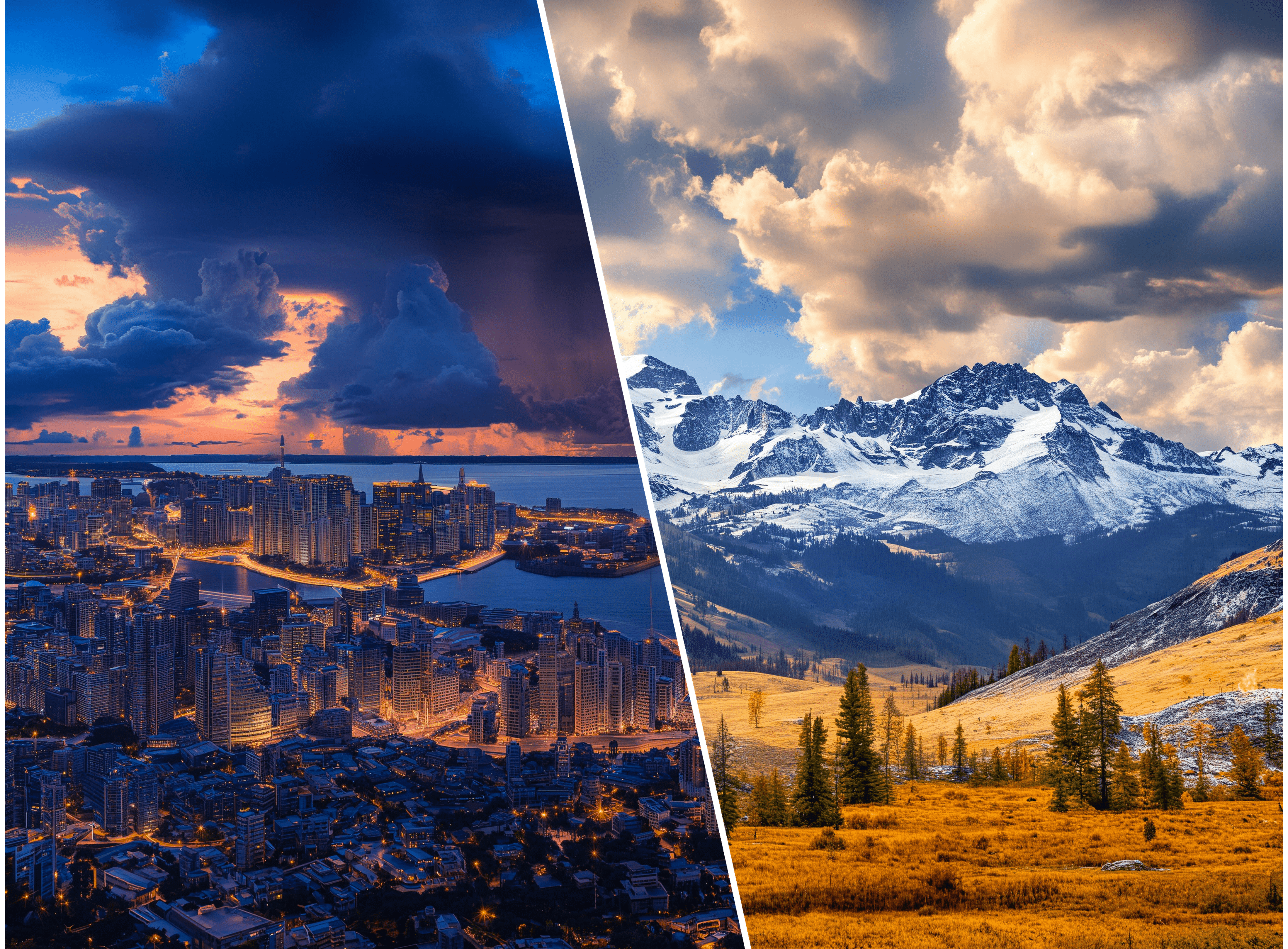}
    \end{minipage}
    \vspace{-2mm}
    \caption{Our UltraHR-100K (\textbf{left}) is a large-scale high-quality dataset for ultra-high-resolution (UHR) image synthesis, featuring a diverse range of categories. Utilizing this dataset enables the generation of high-fidelity UHR images (\textbf{right}). }
    \label{fig:toutu}
\end{figure}
\vspace{-0.08cm}

\begin{abstract}
Ultra-high-resolution (UHR) text-to-image (T2I) generation has seen notable progress. However, two key challenges remain : 1) the absence of a large-scale high-quality UHR T2I dataset, and (2) the neglect of tailored
training strategies for fine-grained detail synthesis in UHR scenarios. To tackle the first challenge, we introduce \textbf{UltraHR-100K}, a high-quality dataset of 100K UHR images with rich captions, offering diverse content and strong visual fidelity. Each image exceeds 3K resolution and is rigorously curated based on detail richness, content complexity, and aesthetic quality. To tackle the second challenge, we propose a frequency-aware post-training method that enhances fine-detail generation in T2I diffusion models. Specifically, we design (i) \textit{Detail-Oriented Timestep Sampling (DOTS)} to focus learning on detail-critical denoising steps, and (ii) \textit{Soft-Weighting Frequency Regularization (SWFR)}, which leverages Discrete Fourier Transform (DFT) to softly constrain frequency components, encouraging high-frequency detail preservation. Extensive experiments on our proposed UltraHR-eval4K benchmarks demonstrate that our approach significantly improves the fine-grained detail quality and overall fidelity of UHR image generation. The code  is available at \href{https://github.com/NJU-PCALab/UltraHR-100k}{here}. %, setting new state-of-the-art performance.
\end{abstract}

\section{Introduction}

Recent advances in text-to-image (T2I) diffusion models have greatly improved image quality and controllability \cite{saharia2022photorealistic, podell2023sdxl, chenpixart, ding2021cogview, peebles2023scalable, esser2024scaling, flux2024, liu2024playground,li2025set,gao2025eraseanything,hu2025exploiting}. However, most existing models are constrained to fixed resolutions (typically 1024×1024), and exhibit noticeable quality degradation and structural artifacts when directly scaled to ultra-high-resolution (UHR) image generation \cite{renultrapixel, bu2025hiflow, du2024demofusion, xie2024sana, zhang2025diffusion, chen2024pixart,zhao2025zero,zhao2024cycle}. This limitation poses a significant barrier for real-world applications that demand fine-grained detail and high visual fidelity, such as digital art, virtual content creation, and commercial design.

Existing solutions to face this challenge can be grouped into two main paradigms: training-free \cite{ bu2025hiflow, du2024demofusion, he2023scalecrafter,huang2024fouriscale, jin2023training, kim2025diffusehigh, zhang2024frecas,zhou2025dragflow,chen2024region} and training-based methods \cite{renultrapixel, xie2024sana, zhang2025diffusion, chen2024pixart,du2025textcrafter}. Training‑free methods attempt to generate UHR images by modifying network architectures \cite{he2023scalecrafter, jin2023training, kim2025diffusehigh} or by adjusting inference schemes \cite{du2024demofusion, huang2024fouriscale}. However, these techniques exhibit excessive smoothing, produce implausible details, and incur prolonged inference times—limitations that severely  hinder their practical deployment~\cite{renultrapixel}.  Fundamentally, training-free methods depend on pre-trained T2I models \cite{podell2023sdxl, chenpixart, peebles2023scalable,flux2024} that were not exposed to UHR data during training, and consequently \textit{lack the inherent capacity to render the fine‑grained, photorealistic details essential} that real-world UHR image synthesis requires.%for true UHR image generation.

\begin{wrapfigure}{r}{7.5 cm}
    \centering
    \includegraphics[width=1\linewidth]{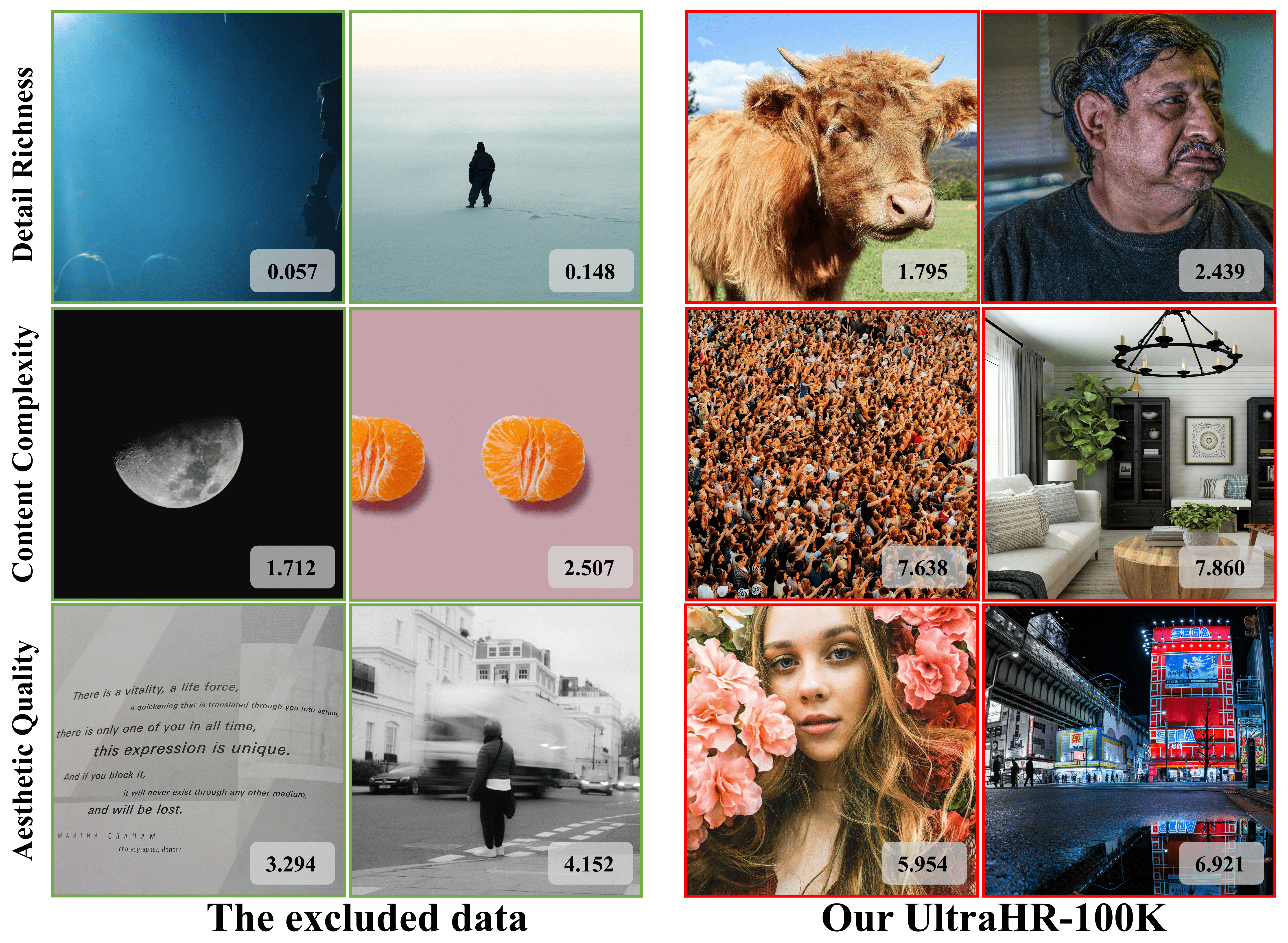}
    \vspace{-4mm}
    \caption{We perform a rigorous selection of UltraHR-100K by evaluating all collected images across three key dimensions: detail richness, content complexity, and aesthetic quality. \textbf{Left:} We present representative low-quality (bad case) examples for each dimension along with their corresponding scores, highlighting the necessity of such filtering. \textbf{Right:} In contrast, our UltraHR-100K exhibit superior texture details, semantic complexity, and aesthetic appeal.}
    \label{fig:UltraHR}
\end{wrapfigure}\vspace{-0.08cm}

Recently, training‑based models for UHR image generation have shown promising results \cite{ xie2024sana, zhang2025diffusion, chen2024pixart}. However, they still face two critical challenges: 1) \textit{The absence of a open‑source, large‑scale high‑quality UHR T2I dataset.} High-fidelity UHR image collection is burdensome due to the scarcity of suitable data. Although Aesthetic-4K \cite{zhang2025diffusion} introduced the first open‑source UHR T2I dataset, it remains limited in both scale (approximately 10K images) and quality (the lack of a rigorous selection criterion), constraining its generalizability and high‑quality generation capabilities in real‑world scenarios. Consequently, constructing a open‑source, large‑scale high‑quality UHR T2I dataset represents both a significant challenge and a critical necessity. 2) \textit{The neglect of tailored training strategies for UHR fine-grained detail synthesis.} Existing models primarily focus on training efficiency to fine-tune pre-trained T2I models \cite{ xie2024sana, chen2024pixart}, overlooking the high‑fidelity detail synthesis. Large-scale pre-training equips T2I models with strong semantic planning abilities, but they struggle to synthesize fine-grained details in the UHR setting \cite{podell2023sdxl, chenpixart, peebles2023scalable,flux2024}. Thus, a detail-oriented training strategy is essential for achieving high-quality UHR image synthesis.%it remains limited in scale (approximately 10K images) and lacks rigorous quality‑filtering criteria, thereby constraining its generalization and high‑quality generation in real‑world scenarios. 

\begin{figure}[t!]
	\centering
	\begin{minipage}[t]{0.36\linewidth}
		\centering
		\includegraphics[width=\linewidth]{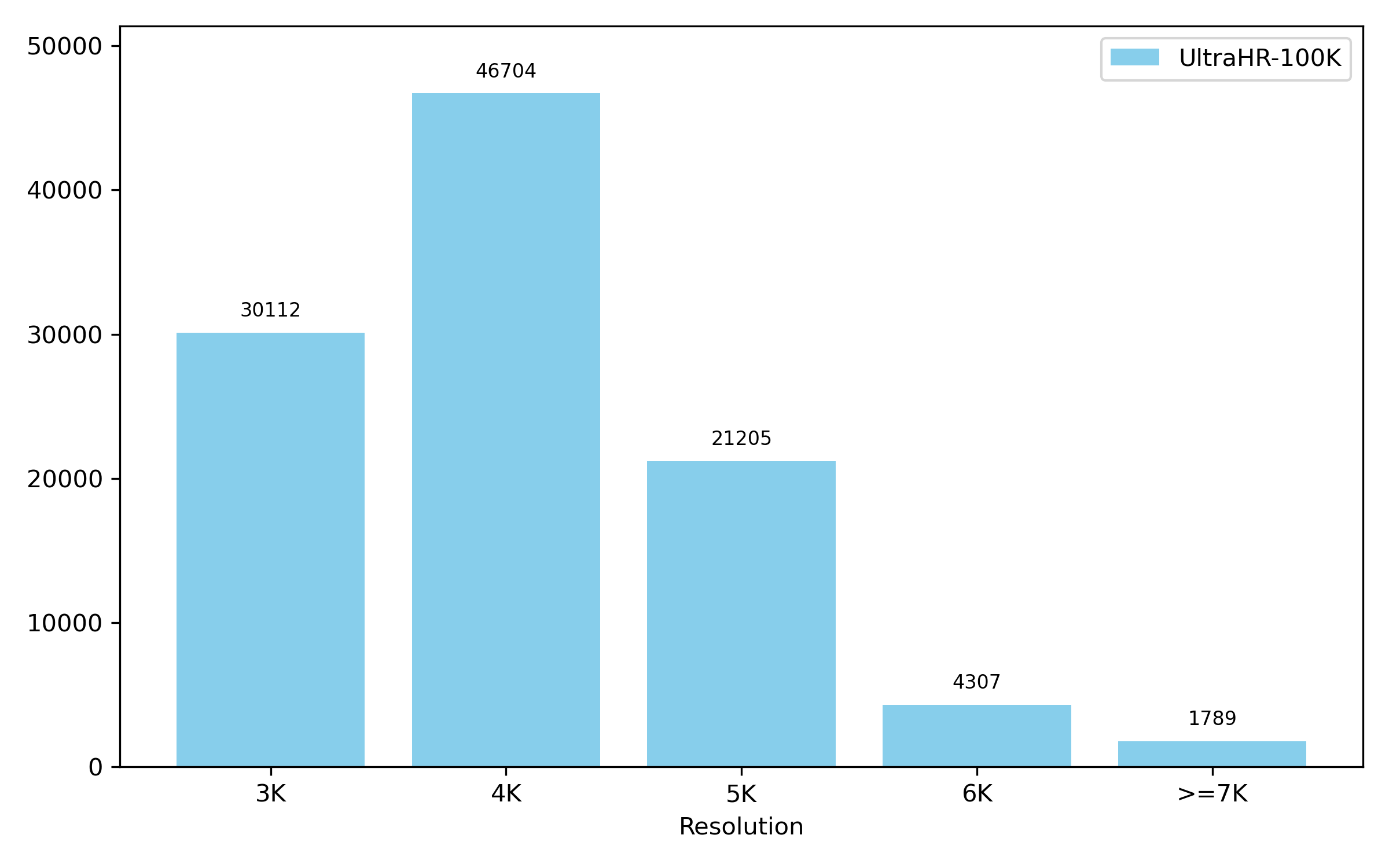}
		\vspace{-2mm}
	\end{minipage}%
	\hfill
	\begin{minipage}[t]{0.25\linewidth}
		\centering
		\includegraphics[width=\linewidth]{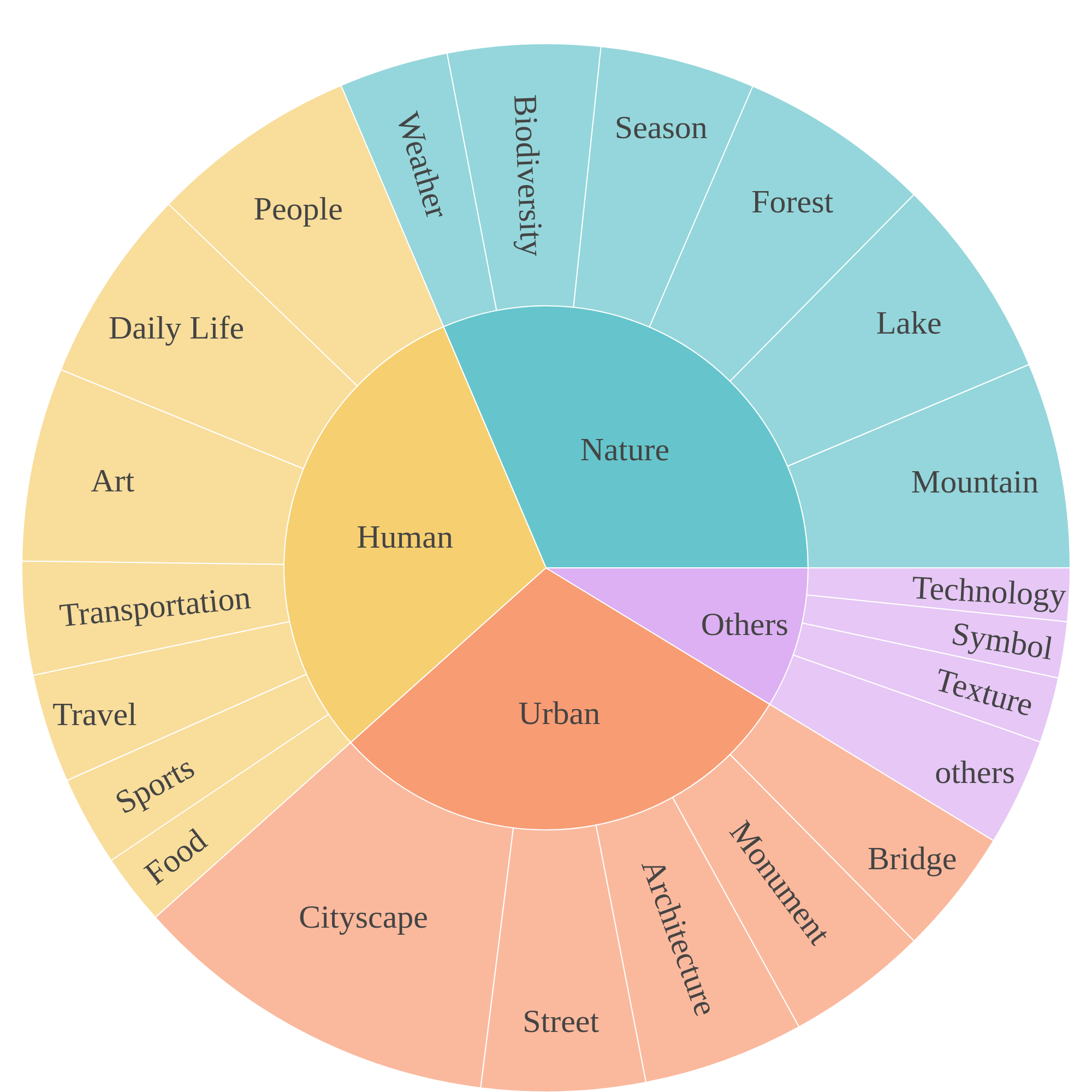}
		\vspace{-2mm}
	\end{minipage}%
	\hfill
	\begin{minipage}[t]{0.36\linewidth}
		\centering
		\includegraphics[width=\linewidth]{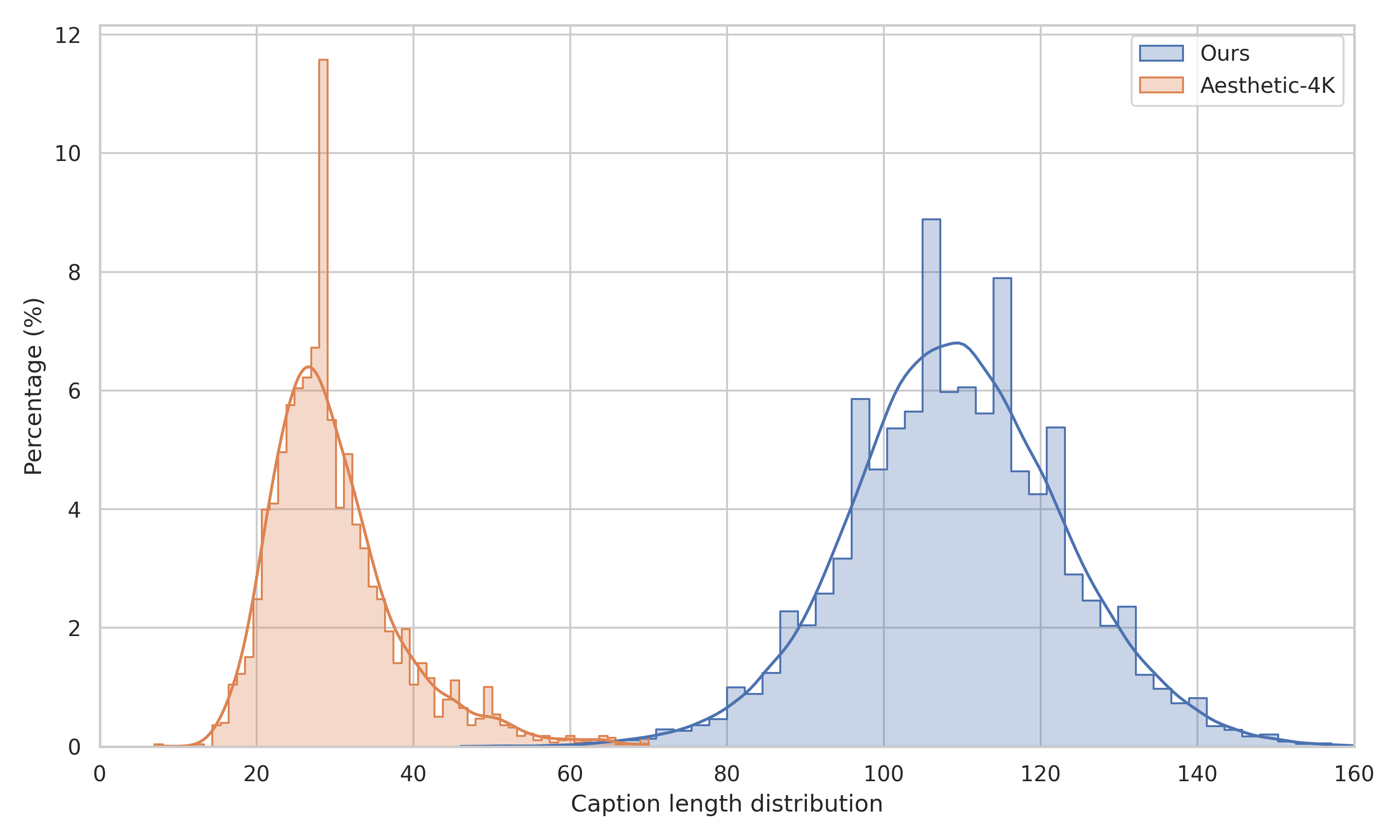}
		\vspace{-2mm}
	\end{minipage}
	\vspace{-2mm}
	\caption{\textbf{Left:} Resolution distribution of our UltraHR-100K. All images have a minimum resolution of 3K, defined as the average of height and width exceeding 3000 pixels. 
		\textbf{Middle:} Image categories across our dataset. \textit{The proportion of each category mirrors its distribution in our dataset.}
		\textbf{Right:} Caption length distribution. Compared to the recent Aesthetic-4K\cite{zhang2025diffusion}, our captions are significantly longer, providing richer semantic supervision.}
	\label{fig:resolution}
\end{figure}
\vspace{-0.08cm}

\textbf{Large-Scale High-Quality Dataset for Tackling Challenge 1:} We construct UltraHR-100K, a large-scale high-quality UHR T2I dataset consisting of 100K UHR images paired with rich textual descriptions. As illustrated in Figure~\ref{fig:toutu}, UltraHR-100K offers the following key advantages: 
1) \textit{Scale and Diversity}: Compared to recent publicly available Aesthetic-4K \cite{zhang2025diffusion}, our UltraHR-100K is approximately 10× larger, featuring 100K images spanning a broad spectrum of categories and visual concepts. 
2) \textit{Higher Quality}: All images in UltraHR-100K are rigorously selected from three key dimensions:  detail richness, content complexity, and aesthetic quality. Notably, the minimum resolution across the proposed dataset exceeds 3K (average of width and height), ensuring high-resolution content, as shown in Figure~\ref{fig:resolution}.
3) \textit{Fine-Grained Captions}: To provide detailed textual annotations for each image, we leverage Gemini 2.0 \cite{gemini}, a powerful commercial vision-language model (VLM), to generate high-quality captions. As shown in Figure~\ref{fig:caption}, our captions are significantly more detailed and semantically rich compared to those in the Aesthetic-4K \cite{zhang2025diffusion}. %offering richer semantic grounding for UHR image generation.

\textbf{Detail-Oriented Training Strategy for Tackling Challenge 2:} To enhance UHR detail synthesis, we propose a frequency-aware post-training method, which consists of detail-oriented timestep sampling (DOTS) and soft-weighting frequency regularization (SWFR). DOTS improves detail synthesis in UHR image generation by directing more training focus to timesteps associated with fine-grained details. Unlike the discrete and block-based decomposition approach used in Diffusion4K \cite{zhang2025diffusion}, which relies on DWT for frequency separation, our SWFR utilizes the continuous spectrum provided by the Discrete Fourier Transform (DFT) to enable more precise frequency control. By applying a soft-weighted constraint across frequency bands, SWFR encourages the model to better reconstruct high-frequency details, without compromising low-frequency structural integrity.

Through the proposed \textit{\textbf{dataset}} and \textit{\textbf{training strategy}}, we can \textit{\textbf{enhance}} the synthesis capability of existing pre-trained T2I models \cite{podell2023sdxl, chenpixart, xie2024sana,flux2024} in UHR image generation, with a particular focus on improving fine-grained detail representation. Furthermore, we construct a large 4K T2I benchmarks,  UltraHR-eval4K (4096 $\times$ 4096), to comprehensively evaluate existing UHR generation models. Extensive experimental results demonstrate the effectiveness of our method.

\section{Related Work}

\begin{table}[t!]

  %\caption{Overview of our data processing pipeline for constructing UltraHR-100K. The first stage involves large-scale data collection and preliminary filtering to ensure a baseline level of visual quality, resulting in an initial subset $S$. The second stage performs three parallel filtering procedures based on \textit{detail richness}, \textit{content complexity}, and \textit{aesthetic quality}, producing subsets $S_G$, $S_E$, and $S_A$ respectively. The final high-quality dataset is obtained by taking the intersection of these subsets. We further employ a strong vision-language model (Gemini 2.0) to annotate each image with long and fine-grained captions.}
  %\caption{Overview of our data processing pipeline for constructing UltraHR-100K. The first stage involves large-scale data collection and preliminary filtering to ensure a baseline level of visual quality. The second stage performs three parallel filtering procedures based on \textit{detail richness}, \textit{content complexity}, and \textit{aesthetic quality}. The final high-quality dataset is obtained by taking the intersection of these subsets. We further employ a strong VLM (Gemini 2.0) to annotate each image with long and fine-grained captions.}
  \caption{Overview of our data processing pipeline. The first stage involves large-scale data collection and preliminary filtering to ensure a baseline level of visual quality. The second stage performs three parallel filtering procedures. The final high-quality dataset is obtained by taking the intersection of these subsets. We further employ a strong VLM (Gemini 2.0) to annotate each image.}

  \label{data_processing}
  \centering
  \resizebox{0.92\linewidth}{!}{
  \begin{tabular}{lllllll}
    \toprule
    Pipeline     &  Tool & Remark    \\
    \midrule
    Data collecting & Python & Get 400K high-resolution images \\
    Preliminary data filtering & Laplacian and Sobel  & Obtain subset $S$ with basic visual quality \\
    \midrule
    Detail richness &  GLCM  & Obtain the set $S_{G}$ with rich fine-grained details \\
    Content complexity & Shannon entropy & Obtain the set $S_{E}$  with complex and diverse content  \\
    Aesthetic score & LAION aesthetic predictor & Get high aesthetic score set $S_{A}$ \\
    \midrule
    The final dataset & Intersection & Obtain intersection: UltraHR-100K = $S_{A} \cap S_{E} \cap S_{G}$ \\
    UHR image caption & Gemini 2.0 & Obtain long and fine-grained descriptions for the images \\
    \bottomrule
  \end{tabular}
  }
\end{table}

%\noindent \textbf{ Text-to-Image Synthesis.}

\subsection{Text-to-Image Synthesis}
Text-to-image (T2I) generation \cite{saharia2022photorealistic, podell2023sdxl, chenpixart, ding2021cogview, peebles2023scalable, esser2024scaling, flux2024, liu2024playground, zhou2024migc, zhangrethinking,zhou2024migc++,zhang2024vocapter,lu2023tf,du2025upgen,nan2024openvid}  has made notable progress owing to the emergence of diffusion-based frameworks \cite{si2024freeu,li2024playground,lu2024mace,zhou20243dis,zhao2024wavelet,zhao2024learning,zhao2024toward,chen2023diffusion,DBLP:journals/corr/abs-2501-05131,DBLP:journals/corr/abs-2504-15376,xie2025star,du2025shift,zhao2025multi}, which exhibit impressive ability in synthesizing visually compelling content from textual descriptions. Early methods such as Denoising Diffusion Probabilistic Models (DDPM) \cite{ho2020denoising} and Denoising Diffusion Implicit Models (DDIM) \cite{song2020denoising} revealed the strength of iterative denoising procedures for producing realistic images. Subsequently, the attention to latent space diffusion \cite{rombach2022high} brought a major breakthrough, significantly lowering training complexity and enhancing scalability \cite{podell2023sdxl}.  More recently, incorporating transformer \cite{ chenpixart, flux2024,  xie2024sana, zhuo2024lumina,zhou2023pyramid}  into diffusion models has further boosted image generation quality.  In this paper, we aim to enhance the generative capability of T2I models in UHR scenarios.

\subsection{Ultra-High-Resolution Image Synthesis}
%\textbf{Ultra-High-Resolution Image Synthesis.} 
UHR image generation plays a crucial role in practical domains such as industry and entertainment \cite{zhang2025diffusion, chai2022any,zhang2025u}. Due to computational constraints, current advanced latent diffusion models typically operate at a maximum resolution of 1024 × 1024 \cite{ podell2023sdxl, chenpixart,  esser2024scaling, flux2024, liu2024playground,zhou2025dreamrenderer,lu2024robust}. However, scaling to 4K resolution significantly increases computational demands, with cost growing quadratically with image size. Several training-free approaches have extended existing latent diffusion models for 4K generation by modifying the inference strategies of diffusion models. \cite{ bu2025hiflow, du2024demofusion,zhang2024catmullrom,he2023scalecrafter,huang2024fouriscale, jin2023training, kim2025diffusehigh,lu2025does}. DiffuseHigh \cite{kim2025diffusehigh} enhances the base-resolution generation by upscaling and subsequently re-denoising it, guided by structural information from the DWT. HiFlow \cite{bu2025hiflow} adopts a cascaded generation paradigm to effectively capture and utilize low-resolution flow characteristics. However, these techniques exhibit excessive smoothing, produce implausible details, and incur prolonged inference time \cite{renultrapixel}.
Pixart-$\sigma$ \cite{chen2024pixart} takes a pioneering step by approaching direct 4K image generation through efficient token compression in DiT. Similarly, Sana \cite{xie2024sana} introduces a cost-effective 4K generation pipeline. Despite these advancements, existing models primarily focus on training efficiency, overlooking the high-fidelity detail synthesis. %Diffusion4K propose a wavelet based fine-tuning approach for direct training with photorealistic 4K images.  

\section{Constructing UltraHR-100K}

\begin{wraptable}{r}{7.2cm}
  \centering
  \caption{Dataset statistical comparisons.}
  \label{dataset_compared}
  \vspace{-3mm}
  \resizebox{0.95\linewidth}{!}{
    \begin{tabular}{c | c |c c }
      \toprule
      Dataset    &  Number &Height & Width  \\
      \midrule
      PixArt-30k \cite{chen2024pixart} & 30,000   & 1,615 & 1,801 \\
      Aesthetic-4K \cite{zhang2025diffusion} & 12,015 & 4,128 & 4,640 \\
      UltraHR-100K & 104,117  & 3,648 & 5,119 \\
      \midrule
      Aesthetic-Eval@4096 \cite{zhang2025diffusion} & 195 & 4,912 & 6,449 \\
      UltraHR-eval4K & 2,000  & 4,912 & 7,175 \\
      \bottomrule
    \end{tabular}
  }
\end{wraptable}

To face the challenge of the lack of high-quality text-image pairs at UHR image generation, we construct a large-scale high-quality dataset named UltraHR-100K. We begin by collecting approximately 400K high-resolution images (with a minimum resolution of 3840×2160) using a custom Python crawler built with Scrapy, sourcing images from the web and various high-resolution imaging devices. \textit{However, high resolution alone does not guarantee high quality. We pose a central question: \textbf{What constitutes a high-quality image for UHR image generation?}} We argue that beyond resolution, such images should exhibit rich content complexity, fine-grained visual details, and aesthetic appeal. Accordingly, we conduct a rigorous filtering process based on three criteria—\textit{\textbf{content complexity, detail richness, and aesthetic quality}}—to curate a 100K-level T2I dataset that meet these standards. The proposed UltraHR-100K provides a reliable foundation for training and evaluating models in high-fidelity UHR image generation. The data processing pipeline is provided in Table~\ref{data_processing}.

\paragraph{Preliminary Data Filtering.}
High-resolution images scraped from the web often suffer from blur, noise, or lack of texture, which can significantly degrade image quality. To eliminate such artifacts, we apply a two-stage low-level quality filter. First, we compute the Laplacian variance to assess image sharpness and discard samples below a blur threshold. Second, we apply the Sobel operator to measure edge density, removing overly flat or textureless images. This process yields a cleaned subset $S$ with sufficient basic visual quality.

\paragraph{Detail Richness.}
Fine-grained details are essential for training generative models to preserve high-frequency content. To quantify the aspect, we compute Gray-Level Co-occurrence Matrix (GLCM) score, including contrast, entropy, and correlation across multiple directions. These metrics capture spatial pixel relationships indicative of texture complexity. We then select the top 50\% of images from $S$ with the highest aggregated GLCM scores, resulting in subset $S_G$.

\paragraph{Content Complexity.}
Visually complex images and diverse spatial structures are more valuable for guiding generation models to achieve rich content. We use Shannon entropy as a proxy to measure the content complexity of each image. Images with higher entropy tend to contain more varied pixel intensities. From subset $S$, we retain the top 50\% highest-entropy images to construct subset $S_E$.

\paragraph{Aesthetic Quality.}
Aesthetic appeal is an important factor in image realism and human preference. To incorporate this dimension, we adopt the LAION Aesthetic Predictor \cite{schuhmann2022laion}, a neural network trained to estimate perceptual quality. It outputs a scalar score reflecting visual composition, color harmony, and overall appeal. We rank all images in $S$ by their aesthetic scores and retain the top 50\% to form subset $S_A$, consisting of the most visually pleasing samples.

\paragraph{UltraHR-100K.}  
To ensure that the final dataset consists of high-quality UHR images with \textit{diverse content, rich textures, and strong aesthetic appeal}, we take the intersection of the three selected subsets. Specifically, the final dataset  is defined as:
\begin{equation}
\text{UltraHR-100K} = S_G \cap S_E \cap S_A
\end{equation}
This intersection guarantees that each image in UltraHR-100K simultaneously meets high standards in detail richness, content complexity, and aesthetic quality, as shown in Figure~\ref{fig:UltraHR}. In addition, we construct a evaluation subset from our dataset—\textbf{UltraHR-eval4K}—containing 2,000  images. Table~\ref{dataset_compared} compares our UltraHR-100K with Aesthetic-4K \cite{zhang2025diffusion} and PixArt-30K \cite{chen2024pixart}. These statistics highlight that UltraHR-100K not only improves dataset scale, but also provides more extensive spatial content.

\begin{figure}[t!]
    \centering
    \includegraphics[width=1\linewidth]{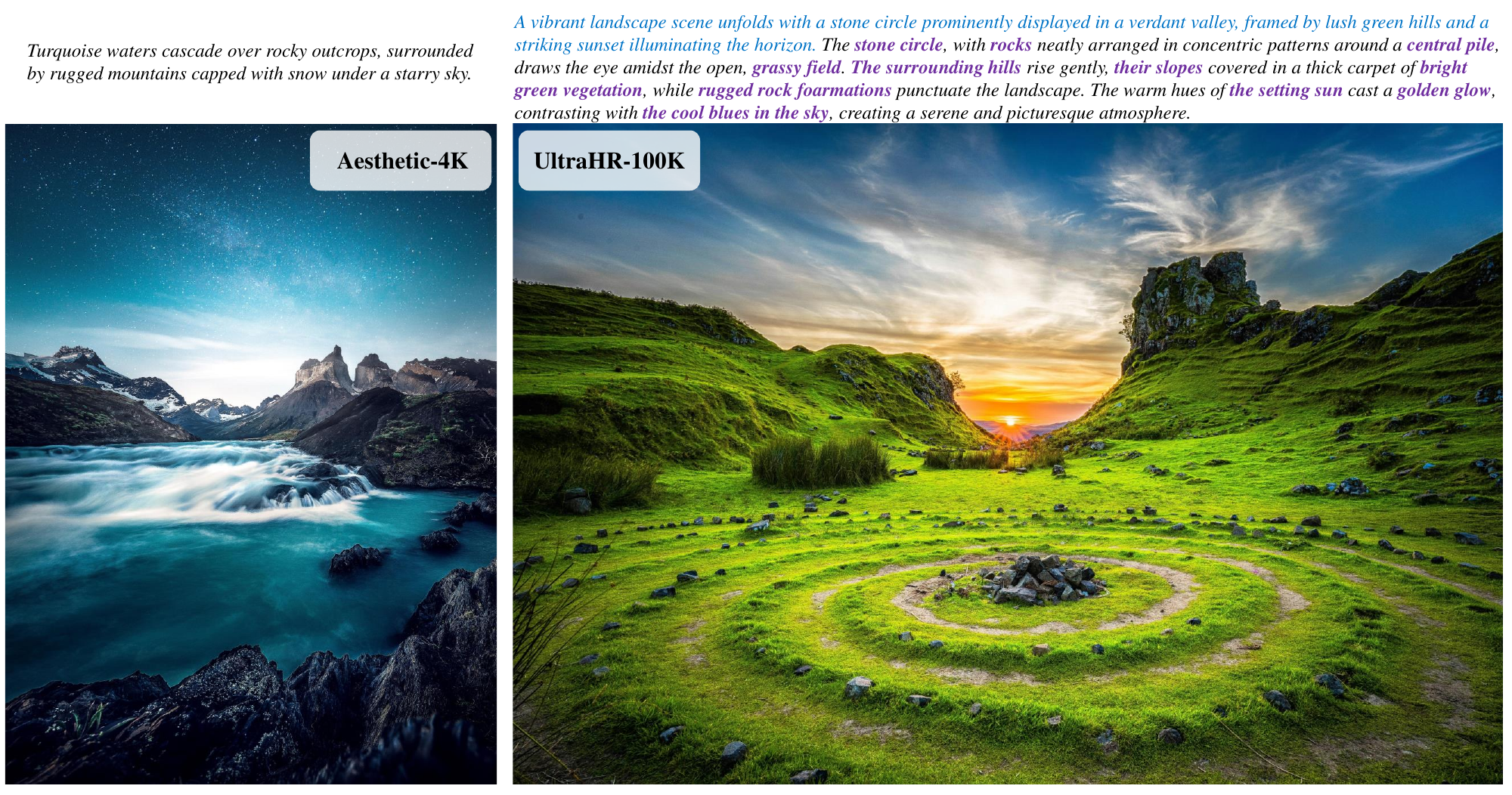}
    \vspace{-4mm}
    \caption{Comparison between our UltraHR-100K and Aesthetic-4K\cite{zhang2025diffusion}. Captions in our UltraHR-100K provide more expressive descriptions, encompassing not only \textcolor{myBlue}{global summaries} of the image content but also \textcolor{myPurple}{rich details} that enhance semantic alignment.}
    \label{fig:caption}
\end{figure}\vspace{-0.08cm}

\paragraph{UHR Image Caption.}

\begin{wrapfigure}{r}{7.5cm}
	\includegraphics[width=0.95\linewidth]{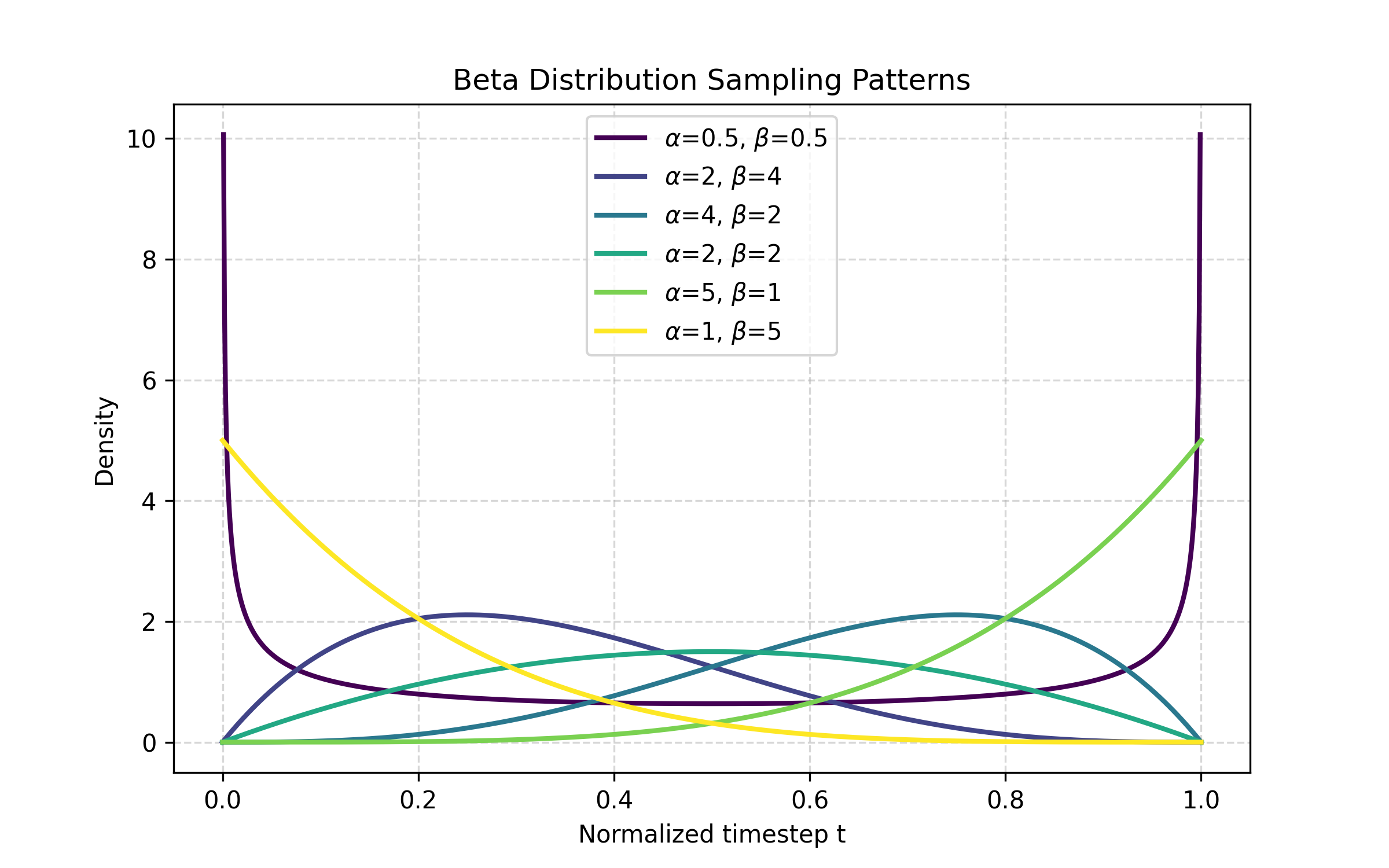}\
	\vspace{-2mm}
	\caption{Relation between weighting ratio and timesteps with beta sampling strategy.}
	% \caption{\textbf{Beta distribution}. }
	\label{fig:beta}
\end{wrapfigure}
UHR images typically contain significantly more visual information than standard-resolution images, making them inherently more semantically complex. However, existing datasets \cite{zhang2025diffusion, schuhmann2022laion} often provide only short captions, limiting the semantic expressiveness of generative models. To address this issue, we leverage Gemini 2.0 \cite{gemini}, a state-of-the-art commercial vision-language model (VLM), to generate rich and detailed captions for our dataset. As illustrated in Figures~\ref{fig:resolution} and~\ref{fig:caption}, our captions are not only substantially longer but also encompass both global summaries and fine-grained descriptions, enhancing alignment with complex image content.

\section{Frequency-Aware Post-Training}

Pretrained T2I models, trained on large-scale datasets, exhibit strong capabilities in semantic and content planning. However, they often struggle to synthesize fine-grained details when extended to UHR scenarios \cite{renultrapixel, xie2024sana}. In this work, we focus on enhancing the detail synthesis ability of pretrained T2I models through tailored post-training strategies. To this end, we propose a frequency-aware post-training method (FAPT). Specifically, FAPT consists of two parts: detail-oriented timestep sampling (DOTS) and soft-weighting frequency regularization (SWFR). DOTS improves detail synthesis in UHR image generation by directing more training focus to timesteps associated with fine-grained details. Meanwhile, SWFR imposes a soft-weighted constraint across the frequency spectrum, guiding the model to better preserve and reconstruct high-frequency details.

\subsection{Detail-Oriented Timestep Sampling}

\noindent \textbf{Motivation.} Existing study \cite{yitowards} have validated the observation that the overall image structure (low-frequency signals) is largely reconstructed in the early denoising steps, while fine-grained details (high-frequency signals) are progressively synthesized in the later stages of the denoising process. This insight motivates us to design a sampling strategy that emphasizes the later stages of the denoising process, aiming to enhance the learning of fine-grained details during post-training stage. %for UHR T2I models.

\noindent \textbf{DOTS.} To achieve this target, we adopt a beta sampling strategy, which provides a simple yet flexible mechanism to bias the sampling distribution over denoising timesteps, as shown in Figure~\ref{fig:beta}.  %As illustrated in Figure \ref{fig:method}, compared to the popular uniform and logit-normal sampling methods, beta sampling better biases the sampling distribution. 
Specifically, we first draw a timestep $t \in (0, 1)$ from a Beta distribution parameterized by shape parameters $\alpha$ and $\beta$:
\begin{equation}
t \sim \mathrm{Beta}(\alpha, \beta).
\end{equation}
The Beta distribution yields a rich family of unimodal or skewed distributions over the interval $(0, 1)$, and its probability density function is given by:
\begin{equation}
\pi_{\mathrm{beta}}(t; \alpha, \beta) = \frac{1}{\mathrm{B}(\alpha, \beta)} t^{\alpha - 1} (1 - t)^{\beta - 1},
\label{eq:beta_pdf}
\end{equation}
where $\mathrm{B}(\alpha, \beta) = \frac{\Gamma(\alpha)\Gamma(\beta)}{\Gamma(\alpha + \beta)}$ is the Beta function. By adjusting $\alpha$ and $\beta$, we can control the bias of the sampling distribution.  This sampling mechanism naturally supports adaptive emphasis in training: by emphasizing later denoising timesteps, we can guide the model to focus on  high-frequency details. %critical for UHR image generation.  

\subsection{Soft-Weighting Frequency Regularization}

\noindent \textbf{Motivation.} Large pre-trained T2I models \cite{podell2023sdxl, chenpixart, peebles2023scalable, flux2024} demonstrate strong semantic planning from diverse data exposure but struggle with fine-grained detail synthesis in UHR scenarios. Existing UHR T2I models focus mainly on training efficiency \cite{xie2024sana, chen2024pixart}, often neglecting high-fidelity detail. Diffusion4K \cite{zhang2025diffusion} introduces DWT-based frequency decomposition to enable 4K training, yet DWT yields coarse and discontinuous frequency separation, limiting its effectiveness for UHR modeling. To overcome this, we adopt DFT-based decomposition, which provides finer, globally coherent frequency representations better suited for capturing fine-scale structures in high-resolution synthesis.

\noindent \textbf{SWFR.} To enhance fine-scale fidelity in UHR image synthesis, we introduce a soft-weighting frequency regularization that complements the standard diffusion loss by explicitly supervising frequency consistency, with an emphasis on high-frequency components. Formally, consider the standard diffusion process:

\begin{equation}
\bm{z}_t = \alpha_t \cdot \bm{x}_0 + \sigma_t \cdot \bm{\epsilon}, \label{eq:diffusion}
\end{equation}

where $\bm{x}_0$ denotes the data distribution, $\bm{\epsilon}$ is sampled from standard normal distribution, and $\alpha_t$, $\sigma_t$ are known coefficients in the diffusion formulation. Recent T2I models \cite{ peebles2023scalable, flux2024, xie2024sana} adopt rectified flows to predict velocity $\bm{v}$, with the objective as follows: %, with the objective as follows:%The baseline loss is defined either as noise prediction:
\begin{equation}
 \bm{v}_\Theta(\bm{z}_t, t) = \bm{\epsilon} - \bm{x}_0. \label{eq:rectifiedflow}
\end{equation}
To regularize the model in the frequency domain, we compute the 2D Discrete Fourier Transforms (DFT) of both prediction  $\bm{x}$  and target $\bm{y}$:
\begin{equation}
\hat{\bm{x}} = \mathcal{F}(\bm{x}), \quad \hat{\bm{y}} = \mathcal{F}(\bm{y}),
\end{equation}

where $\mathcal{F}(\cdot)$ denotes the DFT. Let $\bm{x}$ and $\bm{y}$ denote the model prediction (e.g., $\bm{x} = \bm{v}_\Theta(\bm{z}_t, t)$) and target (e.g., $\bm{y} = \bm{\epsilon} - \bm{x}_0$), respectively. We define a frequency-domain regularization term as:
%\begin{equation}
%	\mathcal{L}_{\text{freq}} = \mathbb{E} \left[ \left| w(\bm{r}) \cdot \log(1 + |\hat{\bm{x}}|^2) - w(\bm{r}) \cdot \log(1 + |\hat{\bm{y}}|^2) \right|_2^2 \right], \label{eq:freq_loss}
%\end{equation}
\begin{equation}
\mathcal{L}_{\text{freq}} = \mathbb{E} \left[ \left| w(\bm{r}) \cdot \hat{\bm{x}} - w(\bm{r}) \cdot \hat{\bm{y}} \right|^2 \right], \label{eq:freq_loss}
\end{equation}
where $w(\bm{r})$ is a  frequency soft weighting function designed to boost high-frequency supervision:
\begin{equation}
w(\bm{r}) = 1 + \lambda \cdot \frac{\exp(\gamma \bm{r}) - 1}{\exp(\gamma) - 1}, \quad \bm{r} \in [0, 1],
\end{equation}
and $\bm{r}$ is the normalized distance from the center of the frequency plane.  Hyperparameters $\lambda$ and $\gamma$ control the strength and steepness of high-frequency emphasis, respectively. %Figure \ref{fig:method} demonstrates that our soft weighting mechanism enables fine-grained control over high-frequency signals, where $\lambda = 6$ and $\gamma = 3$. 
Finally, the overall training objective is defined as:
\begin{equation}
\mathcal{L}_{\text{total}} = \mathcal{L}_{\text{diff}} + \lambda_{\text{freq}} \cdot \mathcal{L}_{\text{freq}}, \label{eq:total_loss}
\end{equation}
where $\mathcal{L}_{\text{diff}}$ denotes the diffusion loss, which can be instantiated as velocity prediction loss ($\|\bm{v}_\Theta(\bm{z}_t, t) - (\bm{\epsilon} - \bm{x}_0)\|^2$). $\lambda_{\text{freq}}$ is a balancing coefficient that controls the strength of frequency-domain supervision. This regularization $\mathcal{L}_{\text{freq}}$ encourages the model to maintain consistent spectral power between prediction and target, especially in high-frequency bands.% that are critical for visual fidelity in UHR image synthesis.

 \begin{figure}[t!]
     \centering
    \includegraphics[width=1\linewidth]{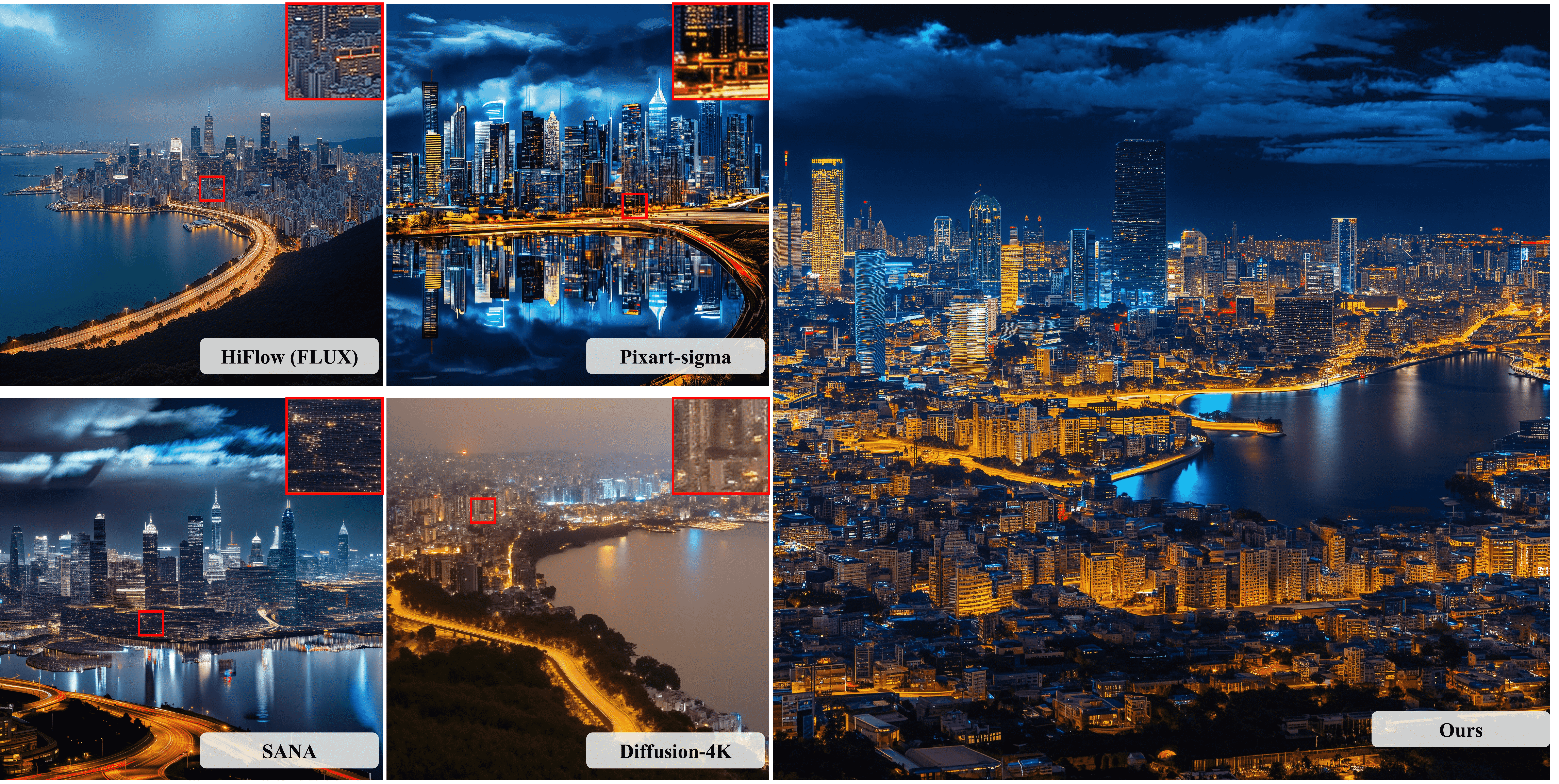}
     \vspace{-4mm}
     \caption{Qualitative comparisons with SOTA methods on our UltraHR-eval4K (4096$ \times$ 4096). Compared with previous works, our method is capable of generating visually complex images with rich semantic content. More visual examples are available in the supplementary materials.}
     \label{fig:visual1}
     	\vspace{-0.5 cm}
 \end{figure}\vspace{-0.08cm}

\begin{figure}[t!]
    \centering
    \includegraphics[width=1\linewidth]{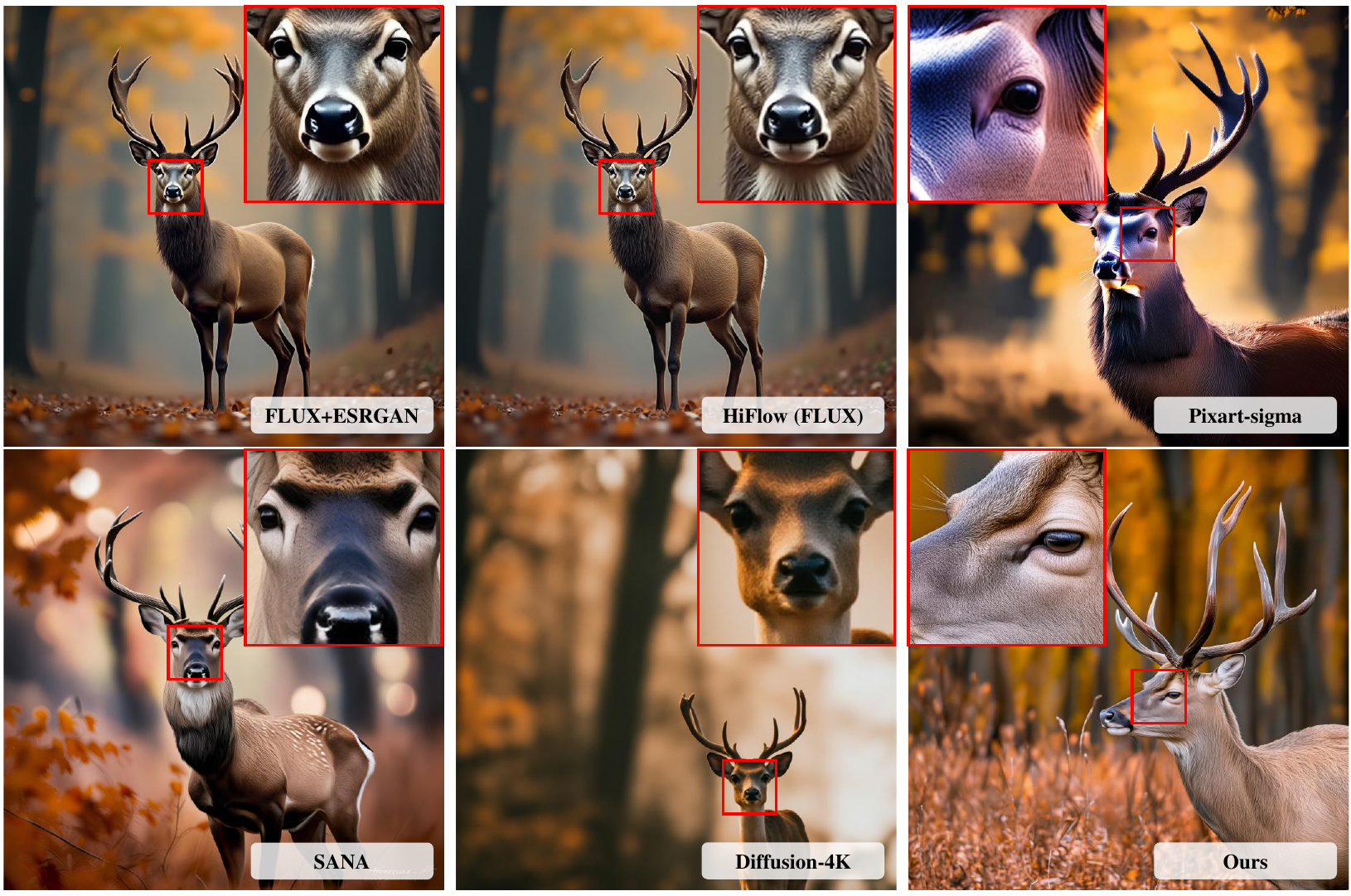}
    \vspace{-4mm}
    \caption{Qualitative comparisons with SOTA methods on our UltraHR-eval4K (4096$ \times$ 4096). Compared with previous works, our method can generate realistic textures and fine-grained details.}
    \label{fig:visual2}
    	\vspace{-0.5 cm}
\end{figure}\vspace{-0.18cm}

\begin{figure}[t!]
	\centering
	\includegraphics[width=1\linewidth]{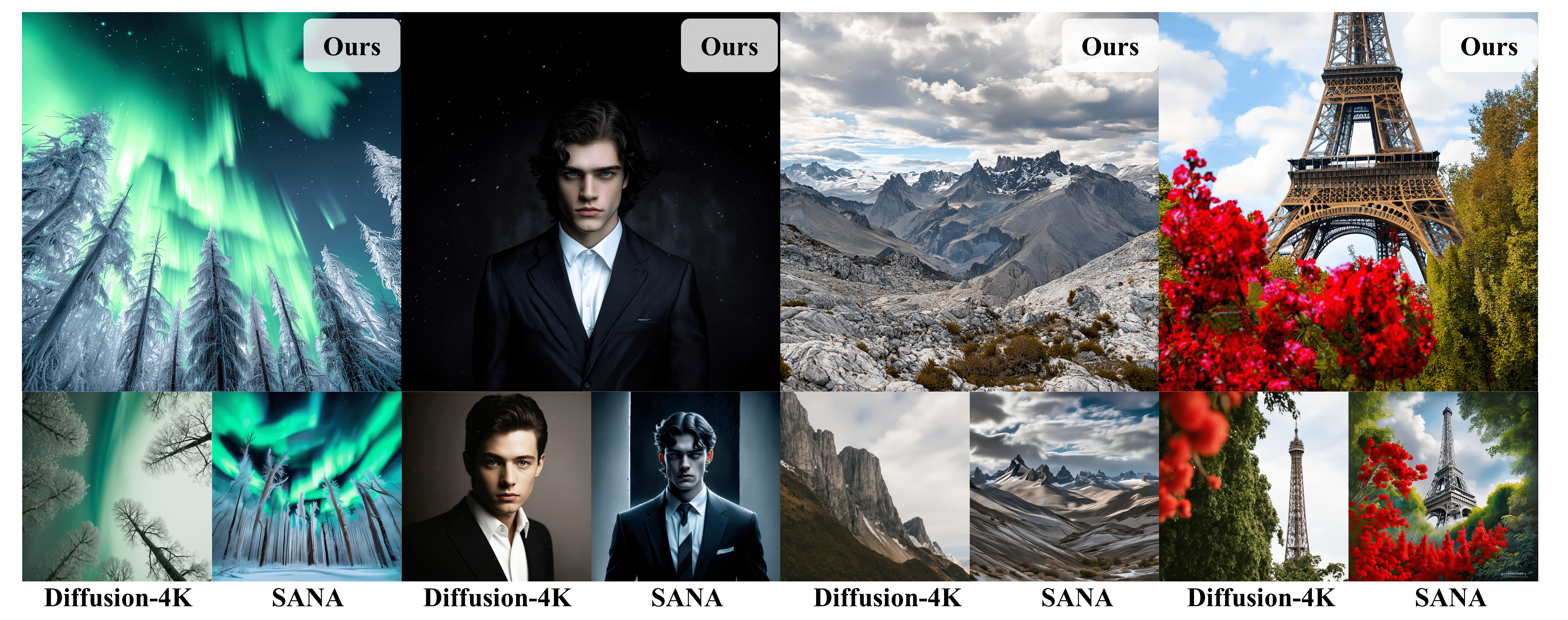}
	\vspace{-4mm}
	\caption{More visual comparisons demonstrate that our method consistently produces high-quality results. Additional and more diverse comparisons can be found in the supplementary material.}
	\vspace{-0.5 cm}
	\label{fig:more}
\end{figure}\vspace{-0.18cm}

\section{Experiments}
\subsection{Implementation Details}
\paragraph{Overall Training Setting.} We adopt a two-stage training strategy. In the first stage, we follow the Logit-Normal Sampling scheme introduced in SD3~\cite{esser2024scaling} and perform fine-tuning on our UltraHR100K dataset, aiming to enhance the  semantic planning capability in UHR generation. In the second stage, we apply our proposed frequency-aware post-training method, which focuses on high-frequency learning to further improve the fine-grained details.  We use the CAMEWrapper~\cite{xie2024sana} optimizer with a constant learning rate of 1e-4, and employ mixed-precision training with a batch size of 24. The first-stage training is conducted for 4K iterations, followed by 8K iterations in the second stage. Due to computational constraints, we conduct training solely on SANA, and all experiments are performed on four H20 GPUs.

\paragraph{Baselines.} To comprehensively evaluate our approach, we conduct extensive comparisons against SOTA methods for UHR image generation, which can be broadly categorized into three groups. The first group consists of powerful T2I models combined with super-resolution technique, BSRGAN~\cite{zhang2021designing}. The second group includes training-free approaches, where we evaluate  FLUX\cite{flux2024}) using corresponding training-free generation methods, I-Max \cite{du2024max} and HiFlow \cite{bu2025hiflow}. Lastly, we compare with leading training-based UHR generation models, including Pixart-$\sigma$ ~\cite{chen2024pixart}, SANA~\cite{xie2024sana}, and Diffusion4K~\cite{zhang2025diffusion}. All baselines are evaluated under their official settings to ensure a fair and consistent comparison.

\paragraph{Evaluation.} We employ several metrics to assess the quality of the generated images, with a particular focus on our evaluation sets, UltraHR-eval4K. To evaluate image-text consistency, we calculate the long CLIP score \cite{zhang2024long} and Fine-Grained (FG) CLIP score \cite{xie2025fg}. Additionally, the Fréchet Inception Distance (FID) \cite{heusel2017gans} and Inception Score (IS) \cite{salimans2016improved} are computed to evaluate the overall image quality of the generated images. Following previous works \cite{bu2025hiflow, renultrapixel}, we compute the FID-patch and IS-patch to evaluate the local quality and details of the images, which are based on local image patches. These metrics provide a comprehensive evaluation of the 
 overall quality, detail retention in the generated images.

%\vspace{-5mm}
\subsection{Comparison to State-of-the-Art Methods}

\begin{table}[t]
  \centering
  \centering          
  \caption{\textbf{Quantitative comparison with other baselines on our UltraHR-eval4K (4096 $\times$ 4096) benchmark.} The best result is highlighted in \textbf{bold}. %* indicates methods migrated from U-Net architecture. , while the second-best result is \underline{underlined}
  }
   \resizebox{\linewidth}{!}{
   \begin{tabular}{cccccccc}
  \toprule
  \textbf{Method}   & \textbf{FID $\downarrow$}&\textbf{FID$_\text{patch}\downarrow$}&\textbf{IS $\uparrow$}&\textbf{IS$_\text{patch}\uparrow$}&\textbf{CLIP $\uparrow$}&\textbf{FG-CLIP $\uparrow$} \\
  \midrule
  FLUX \cite{flux2024} + BSRGAN~\cite{zhang2021designing}&37.651&43.143&11.773&5.389& 31.45 & 28.02 \\  
 SD3.5 \cite{esser2024scaling} + BSRGAN~\cite{zhang2021designing}& 31.870  & 25.598   &12.780   & 5.456  & 31.75&28.66  \\  
  \midrule
%FreCaS(SDXL)~\cite{zhang2024frecas}&35.611&40.620&12.102&5.884&0.304&6.608&0.710\\

I-Max(FLUX)~\cite{du2024max}&37.667&37.835&11.991&4.391& 31.49 &27.78 \\
HiFlow(FLUX)~\cite{bu2025hiflow}&35.892&38.327&11.767&4.620& 31.52  & 27.75 \\
  \midrule
Pixart-$\sigma$~\cite{chen2024pixart}& 33.171  & 32.198
  &  12.212 & 5.390  &  31.78 &28.65 
 \\
SANA~\cite{xie2024sana}&37.070 &38.795&11.778& \textbf{5.649} &31.70  &28.60\\
%SANA1.5~\cite{xie2025sana}& 37.005  & 59.910  &  12.081  &  4.814    &  0.319 & 6.655 & 0.192 \\
Diffusion4K~\cite{zhang2025diffusion}&  39.857 &  38.515
 &   10.832  &  3.235  &  31.41  &  26.48  \\
  \midrule
 \cellcolor{color3}{\textbf{Ours(UltraHR-100K)}}&\cellcolor{color3}{33.995} &	\cellcolor{color3}{20.932} &	\cellcolor{color3}{12.502} &	\cellcolor{color3}{5.020}&	\cellcolor{color3}\textbf{31.85}& 	\cellcolor{color3}{28.65}\\  
  \midrule
\cellcolor{color3}{\textbf{Ours(UltraHR-100K+FAPT)}}&\cellcolor{color3}\textbf{ 31.748 }&\cellcolor{color3}\textbf{15.795}&  
\cellcolor{color3}\textbf{12.995}&\cellcolor{color3}{5.104}&  
\cellcolor{color3}{31.82} & \cellcolor{color3}\textbf{28.68}\\  
  \bottomrule
\end{tabular}    }       
\label{table:quantitative}  
\vspace{-0.05in}
\end{table}

\begin{table*}[t!] % 使用 table* 环境使其横跨整个页面宽度
	\centering
	%\caption{Left: User study results conducted on our UltraHR-eval4K benchmark, evaluating overall quality, detail fidelity, text-image alignment and user preference across different methods. Right: Quantitative comparison on Aesthetic-Eval@4096. We report FID, FID$_\text{patch}\downarrow$, CLIP, and FG-CLIP metrics. The results demonstrate the superior performance of our model on this public UHD benchmark.}
	\caption{Left: User study results conducted on our UltraHR-eval4K. Right: Quantitative comparison on Aesthetic-Eval@4096. The results demonstrate the superior performance of our method.}
	\label{tab:combined_results}
	
	% --- 左边的表格 ---
	\begin{minipage}[t]{0.47 \textwidth} % 分配给左边表格的宽度
		\centering
		\captionsetup{width=.9\linewidth} % 确保子标题不会太宽
		% 使用 resizebox 确保表格适应 minipage 的宽度
		\resizebox{\linewidth}{!}{
			\begin{tabular}{lcccc}
				\toprule
				\textbf{Method} & \textbf{Overall} & \textbf{Detail} & \textbf{Text-Image} & \textbf{Preference} \\
				& \textbf{Quality} & \textbf{Quality} & \textbf{Alignment} &  \\
				\midrule
				Pixart-$\sigma$~\cite{chen2024pixart} & 14\% & 10\% & 16\% & 18\% \\
				SANA~\cite{xie2024sana}            & 4\%  & 8\%  & 8\%  & 6\%  \\
				Diffusion4K~\cite{zhang2025diffusion}     & 12\% & 4\%  & 6\%  & 6\%  \\
				\midrule
				\textbf{Ours}   & \textbf{70\%} & \textbf{78\%} & \textbf{72\%} & \textbf{70\%} \\
				\bottomrule
			\end{tabular}
		}
		%\subcaption{User study results.}
		\label{tab:user_study}
	\end{minipage}%
	\hfill % 在两个表格之间添加弹性空间
	% --- 右边的表格 ---
	\begin{minipage}[t]{0.52 \textwidth} % 分配给右边表格的宽度
		\centering
		\captionsetup{width=.9\linewidth}
		% 定义小数点对齐的列类型
		\newcolumntype{d}[1]{S[table-format=#1]}
		\resizebox{\linewidth}{!}{
			\begin{tabular}{lcccc}
				\toprule
				\textbf{Method} & \textbf{FID $\downarrow$}&\textbf{FID$_\text{patch}\downarrow$}&\textbf{CLIP $\uparrow$}&\textbf{FG-CLIP $\uparrow$} \\
				\midrule
				Pixart-$\sigma$~\cite{chen2024pixart} & 150.593 & 44.702 & 34.88 & 28.48 \\
				SANA~\cite{xie2024sana}            & 146.027 & 37.031 & 34.62 & 28.61 \\
				Diffusion4K~\cite{zhang2025diffusion}           & 152.790 & 39.729 & 33.99 & 26.06 \\
				\midrule
				\textbf{Ours}   &  \textbf{142.965} &  \textbf{24.008} &  \textbf{35.08} & \textbf{28.64} \\
				\bottomrule
			\end{tabular}
		}
		%\subcaption{Quantitative comparison.}
		\label{tab:quantitative_comparison}
	\end{minipage}
	
\end{table*}

%\begin{wraptable}{r}{9 cm}
\begin{table}[t]
	\centering
	%\caption{Ablation study.}
	\caption{ Ablation study of our key components and data scale. Model A is a baseline using full fine-tuning on our dataset. The comparison between C (trained on a partial dataset) and D (full dataset) validates the effectiveness of large-scale data.}
	\label{ablation}
	%\vspace{-0.05mm}
	\resizebox{0.65\linewidth}{!}{
		\begin{tabular}{c |  c c c  | c c c }
			\toprule
			Model    &  DOTS & SWFR & Dataset & \textbf{FID $\downarrow$}&\textbf{FID$_\text{patch}\downarrow$}& \textbf{CLIP $\uparrow$}  \\
			\midrule
			LoRA & $\times$ & $\times$ &  Full & 35.07 &	35.02&	31.80 \\
			A & $\times$ & $\times$ &  Full & 33.99 &20.93 & \textbf{31.85} \\
			B & $\checkmark$ & $\times$ & Full & 32.57 & 19.95 & 31.79 \\
			\midrule
			%C & $\checkmark$ & $\checkmark$ & Part & 39.58 & 45.86 & 31.42 \\
			C  & $\checkmark$ & $\checkmark$ & Part & 32.75 & 18.42 & 31.81 \\
			\midrule
			D & $\checkmark$  & $\checkmark$ & Full & \textbf{31.74} & \textbf{15.79} & 31.82 \\
			\bottomrule
		\end{tabular}
	}
\vspace{-5mm}
\end{table}
%\end{wraptable}

\paragraph{Quantitative Comparison.} Table \ref{table:quantitative} summarizes the quantitative performance on our UltraHR-eval4K benchmark (4096 $\times$ 4096). Our method consistently achieves superior scores on key perceptual metrics such as FID, FID-patch and IS, indicating its strong capability in generating high-quality images with fine-grained textures. Moreover, our method achieves competitive CLIP scores, reflecting its ability to maintain semantic alignment with the input prompt. Notably, our method yields a substantial improvement in FID$_\text{patch}$, highlighting its effectiveness in synthesizing fine-grained details. This result demonstrates that our proposed approach significantly enhances the detail generation capability of pre-trained T2I models in UHR scenarios.

\paragraph{Qualitative Comparison.} Figure~\ref{fig:visual1} presents qualitative comparisons on UltraHR-eval4K (4096 $\times$ 4096), focusing on the overall semantic richness and spatial layout of the generated images. While existing SOTA methods struggle to produce coherent and content-rich scenes at such ultra-high resolution, our method demonstrates a strong capability in generating visually complex images with diverse and semantically meaningful elements. This highlights our model’s superior capacity for global spatial reasoning and semantic planning in large-scale synthesis.
In Figure~\ref{fig:visual2}, we further compare fine-grained textures and local details. Our method produces sharper structures and more realistic textures, faithfully preserving high-frequency information that other methods tend to miss or oversmooth. These results collectively demonstrate the effectiveness of our proposes dataset and method in enhancing both the global semantics and local fidelity for ultra-high-resolution text-to-image generation. Figure~\ref{fig:more} presents more visual comparisons.

\paragraph{User Study.} As shown in Table \ref{tab:combined_results}, we conducted a user study with 5 volunteers evaluating 50 randomly selected cases. Images were rated on overall quality, detail quality, text-image alignment and preference. The results demonstrate the superiority of our method across all aspects.

\paragraph{Comparisons on Public Benchmark.} We conduct a quantitative comparison on the publicly available Aesthetic-4K benchmark, specifically the Aesthetic-Eval@4096 subset, as reported in Table \ref{tab:combined_results}.  This evaluation set contains 195 image-text pairs, where all images have a short side greater than 4096 pixels. Due to the limited number of samples, the reported FID scores are relatively high. Nonetheless, the results clearly demonstrate the superior performance of our method, supporting its robustness and generalizability beyond our proposed benchmark.

%\vspace{-5mm}
\subsection{ Ablation Study}
We conduct a comprehensive ablation study to validate the effectiveness of our proposed training strategy and the importance of large-scale data. As shown in Table~\ref{ablation}, Model A serves as the baseline without our proposed DOTS and SWFR. Model B introduces DOTS, resulting in consistent improvements in both FID and patch-level FID, demonstrating its effectiveness in guiding the sampling process. Further incorporating SWFR (Model D) yields substantial improvements, particularly in patch-level FID, confirming that our proposed regularization enhances the detail synthesis capability of T2I models. To evaluate the impact of training data scale, we compare Model C and Model D. Model C is trained with a randomly sampled 15K subset of our UltraHR-100K using the same training strategy. The performance drop compared to Model D clearly highlights the importance of large-scale UHR data in achieving high-fidelity and semantically aligned image generation.

%\begin{table}[h!]
	\begin{wraptable}{r}{7 cm}
	\centering
	\caption{Analysis for DOTS.}
	\label{tab:dots_ablation}

	\newcolumntype{d}[1]{S[table-format=#1]}
	\resizebox{0.95 \linewidth}{!}{
	\begin{tabular}{c  c c c}
		\toprule
		\textbf{Method} & {\textbf{FID}} & {\textbf{FID\_patch}} & {\textbf{CLIP}} \\
		\midrule
		($\alpha = 3, \beta = 4$) & 33.196 & 22.143 & 31.83 \\
		($\alpha = 1, \beta = 4$) & 33.727 & 25.095 & 31.79 \\
		($\alpha = 2, \beta = 5$) & 33.874 & 23.850 & 31.82 \\
		($\alpha = 2, \beta = 3$) & 33.638 & 24.638 & \textbf{31.84} \\
		\midrule % 添加一条线以突出我们的方法
		\textbf{($\alpha = 2, \beta = 4$)} & \textbf{31.748} &  \textbf{15.795} &  31.82 \\
		\bottomrule
	\end{tabular} }
\end{wraptable}
%\end{table}

\paragraph{Analysis for DOTS.} The DOTS module employs a Beta($\alpha$, $\beta$) distribution to guide timestep sampling, where $\alpha$ and $\beta$ control the bias along the denoising trajectory. When $\alpha$ < $\beta$, sampling favors later steps (near $t=0$) that refine high-frequency details; when $\alpha$ > $\beta$, it leans toward early steps (near $t=1$) emphasizing global structure. In our experiments, we set $\alpha=2$, $\beta=4$, biasing sampling toward later steps to better capture fine details crucial for ultra-high-resolution generation. An ablation study (Table \ref{tab:dots_ablation}) varying $\alpha$ and $\beta$ confirms this choice: larger $\alpha$ weakens detail learning, smaller $\alpha$ harms semantic consistency, and overly concentrated or flattened distributions reduce diversity. These results validate $(\alpha=2, \beta=4)$ as a balanced and effective configuration. %The DOTS module utilizes a Beta($\alpha$, $\beta$) distribution to guide the timestep sampling process. The parameters $\alpha$ and $\beta$ are used to control the sampling bias toward different phases of the denoising trajectory: When $\alpha$ < $\beta$, the distribution favors later denoising steps (closer to t = 0), which tend to focus on high-frequency detail refinement. When $\alpha$ > $\beta$, the sampling shifts toward early denoising steps (closer to t = 1), which contributes more to global semantic structure. In our main experiments, we set $\alpha$ = 2 and $\beta$ = 4, intentionally biasing the sampling toward later timesteps to better capture fine details, which are critical for ultra-high-resolution image generation. To further investigate the sensitivity of $\alpha$ and $\beta$, we conducted an ablation study, as shown in Table 1. Below is a summary of our findings: ($\alpha$ = 3, $\beta$ =4): Increasing $\alpha$ shifts the distribution rightward, leading to more early-step sampling. This negatively impacts high-frequency learning, as the model receives less supervision in later steps. ($\alpha$ =1, $\beta$ =4): Decreasing $\alpha$ moves the distribution leftward, favoring later-step sampling. This can reduce semantic fidelity, as the model may not learn sufficient global structure. ($\alpha$ =2, $\beta$ =5): Increases the distribution’s peak near the center, causing sampling to concentrate too narrowly, limiting diversity in supervision. ($\alpha$ =2, $\beta$ =3): Lowers the distribution peak, increasing sampling across both early and late steps. While more balanced, this can dilute the benefits of our targeted sampling strategy. These results support our choice of ($\alpha$ = 2, $\beta$ = 4) as a balanced and effective configuration that encourages detail learning without sacrificing semantic structure. We will include these discussions and the corresponding table in the revised paper to improve clarity.

\vspace{-1mm}
\section{Conclusion}
%In this paper, we present UltraHR-100K, a curated dataset of 100K UHR images with rich textual annotations. Each image is carefully selected to ensure high levels of detail, visual complexity, and aesthetic appeal. Moreover, we introduce a frequency-aware post-training method, which includes: (i) Detail-Oriented Timestep Sampling (DOTS), which allocates more training focus to denoising steps critical for detail reconstruction, and (ii) Soft-Weighting Frequency Regularization (SWFR), which employs the DFT to gently guide the model toward preserving high-frequency components. %Experiments on our proposed UltraHR-eval4K benchmark confirm that our approach significantly boosts both the visual fidelity and fine-detail accuracy of UHR image synthesis.

In this paper, we present UltraHR-100K, a curated dataset of 100K UHR images with rich textual annotations. Each image is carefully selected to ensure high levels of detail, visual complexity, and aesthetic appeal. Moreover, we introduce a frequency-aware post-training method, which includes: (i) Detail-Oriented Timestep Sampling (DOTS), and (ii) Soft-Weighting Frequency Regularization (SWFR). Experiments on our proposed UltraHR-eval4K benchmark confirm that our approach significantly boosts both the visual fidelity and fine-detail accuracy of UHR image synthesis.

\paragraph{Limitations and future works.} Our main limitations lie in two aspects. First, while the proposed frequency-aware post-training strategy effectively enhances fine-detail synthesis, it introduces a slight degradation in text–image alignment, as shown in Table \ref{ablation}. Second, our dataset currently contains a relatively limited amount of portrait data, which constrains the improvement in ultra-high-resolution (UHR) portrait generation, as illustrated in Figure \ref{fig:more}. In future work, we plan to develop more balanced training strategies to alleviate the alignment issue and expand our dataset with additional high-quality UHR portrait images to further improve performance in portrait synthesis.%Although our frequency-awre post-training strategy enhances detail synthesis, it slightly compromises text-image alignment, as shown in Table \ref{ablation}. In future work, we will explore more effective training strategies to mitigate this issue.

\paragraph{Acknowledgments.} This work was supported by Natural Science Foundation of China: No. 62406135, Natural Science Foundation of Jiangsu Province: BK20241198, and Gusu Innovation and Entrepreneur Leading Talents: No. ZXL2024362.

\newpage
\bibliography{reference}

\begin{thebibliography}{10}

\bibitem{saharia2022photorealistic}
Chitwan Saharia, William Chan, Saurabh Saxena, Lala Li, Jay Whang, Emily~L
  Denton, Kamyar Ghasemipour, Raphael Gontijo~Lopes, Burcu Karagol~Ayan, Tim
  Salimans, et~al.
\newblock Photorealistic text-to-image diffusion models with deep language
  understanding.
\newblock {\em Advances in neural information processing systems},
  35:36479--36494, 2022.

\bibitem{podell2023sdxl}
Dustin Podell, Zion English, Kyle Lacey, Andreas Blattmann, Tim Dockhorn, Jonas
  M{\"u}ller, Joe Penna, and Robin Rombach.
\newblock Sdxl: Improving latent diffusion models for high-resolution image
  synthesis.
\newblock {\em arXiv preprint arXiv:2307.01952}, 2023.

\bibitem{chenpixart}
Junsong Chen, YU~Jincheng, GE~Chongjian, Lewei Yao, Enze Xie, Zhongdao Wang,
  James Kwok, Ping Luo, Huchuan Lu, and Zhenguo Li.
\newblock Pixart-$\alpha$: Fast training of diffusion transformer for
  photorealistic text-to-image synthesis.
\newblock In {\em The Twelfth International Conference on Learning
  Representations}.

\bibitem{ding2021cogview}
Ming Ding, Zhuoyi Yang, Wenyi Hong, Wendi Zheng, Chang Zhou, Da~Yin, Junyang
  Lin, Xu~Zou, Zhou Shao, Hongxia Yang, et~al.
\newblock Cogview: Mastering text-to-image generation via transformers.
\newblock {\em Advances in neural information processing systems},
  34:19822--19835, 2021.

\bibitem{peebles2023scalable}
William Peebles and Saining Xie.
\newblock Scalable diffusion models with transformers.
\newblock In {\em Proceedings of the IEEE/CVF international conference on
  computer vision}, pages 4195--4205, 2023.

\bibitem{esser2024scaling}
Patrick Esser, Sumith Kulal, Andreas Blattmann, Rahim Entezari, Jonas
  M{\"u}ller, Harry Saini, Yam Levi, Dominik Lorenz, Axel Sauer, Frederic
  Boesel, et~al.
\newblock Scaling rectified flow transformers for high-resolution image
  synthesis.
\newblock In {\em Forty-first international conference on machine learning},
  2024.

\bibitem{flux2024}
{Black-Forest Labs}.
\newblock {Flux}.
\newblock \url{https://huggingface.co/black-forest-labs/FLUX.1-dev}, 2024.

\bibitem{liu2024playground}
Bingchen Liu, Ehsan Akhgari, Alexander Visheratin, Aleks Kamko, Linmiao Xu,
  Shivam Shrirao, Chase Lambert, Joao Souza, Suhail Doshi, and Daiqing Li.
\newblock Playground v3: Improving text-to-image alignment with deep-fusion
  large language models.
\newblock {\em arXiv preprint arXiv:2409.10695}, 2024.

\bibitem{li2025set}
Leyang Li, Shilin Lu, Yan Ren, and Adams Wai-Kin Kong.
\newblock Set you straight: Auto-steering denoising trajectories to sidestep
  unwanted concepts.
\newblock {\em arXiv preprint arXiv:2504.12782}, 2025.

\bibitem{gao2025eraseanything}
Daiheng Gao, Shilin Lu, Wenbo Zhou, Jiaming Chu, Jie Zhang, Mengxi Jia, Bang
  Zhang, Zhaoxin Fan, and Weiming Zhang.
\newblock Eraseanything: Enabling concept erasure in rectified flow
  transformers.
\newblock In {\em Forty-second International Conference on Machine Learning},
  2025.

\bibitem{hu2025exploiting}
Xiantao Hu, Ying Tai, Xu~Zhao, Chen Zhao, Zhenyu Zhang, Jun Li, Bineng Zhong,
  and Jian Yang.
\newblock Exploiting multimodal spatial-temporal patterns for video object
  tracking.
\newblock In {\em Proceedings of the AAAI Conference on Artificial
  Intelligence}, volume~39, pages 3581--3589, 2025.

\bibitem{renultrapixel}
Jingjing Ren, Wenbo Li, Haoyu Chen, Renjing Pei, Bin Shao, Yong Guo, Long Peng,
  Fenglong Song, and Lei Zhu.
\newblock Ultrapixel: Advancing ultra high-resolution image synthesis to new
  peaks.
\newblock In {\em The Thirty-eighth Annual Conference on Neural Information
  Processing Systems}.

\bibitem{bu2025hiflow}
Jiazi Bu, Pengyang Ling, Yujie Zhou, Pan Zhang, Tong Wu, Xiaoyi Dong, Yuhang
  Zang, Yuhang Cao, Dahua Lin, and Jiaqi Wang.
\newblock Hiflow: Training-free high-resolution image generation with
  flow-aligned guidance.
\newblock {\em arXiv preprint arXiv:2504.06232}, 2025.

\bibitem{du2024demofusion}
Ruoyi Du, Dongliang Chang, Timothy Hospedales, Yi-Zhe Song, and Zhanyu Ma.
\newblock Demofusion: Democratising high-resolution image generation with no
  \$\$\$.
\newblock In {\em Proceedings of the IEEE/CVF Conference on Computer Vision and
  Pattern Recognition}, pages 6159--6168, 2024.

\bibitem{xie2024sana}
Enze Xie, Junsong Chen, Junyu Chen, Han Cai, Haotian Tang, Yujun Lin, Zhekai
  Zhang, Muyang Li, Ligeng Zhu, Yao Lu, et~al.
\newblock Sana: Efficient high-resolution image synthesis with linear diffusion
  transformers.
\newblock {\em arXiv preprint arXiv:2410.10629}, 2024.

\bibitem{zhang2025diffusion}
Jinjin Zhang, Qiuyu Huang, Junjie Liu, Xiefan Guo, and Di~Huang.
\newblock Diffusion-4k: Ultra-high-resolution image synthesis with latent
  diffusion models.
\newblock In {\em Proceedings of the IEEE/CVF Conference on Computer Vision and
  Pattern Recognition}, 2025.

\bibitem{chen2024pixart}
Junsong Chen, Chongjian Ge, Enze Xie, Yue Wu, Lewei Yao, Xiaozhe Ren, Zhongdao
  Wang, Ping Luo, Huchuan Lu, and Zhenguo Li.
\newblock Pixart-$\sigma$: Weak-to-strong training of diffusion transformer for
  4k text-to-image generation.
\newblock In {\em European Conference on Computer Vision}, pages 74--91.
  Springer, 2024.

\bibitem{zhao2025zero}
Chen Zhao, Zhizhou Chen, Yunzhe Xu, Enxuan Gu, Jian Li, Zili Yi, Qian Wang,
  Jian Yang, and Ying Tai.
\newblock From zero to detail: Deconstructing ultra-high-definition image
  restoration from progressive spectral perspective.
\newblock In {\em Proceedings of the Computer Vision and Pattern Recognition
  Conference}, pages 17935--17946, 2025.

\bibitem{zhao2024cycle}
Chen Zhao, Weiling Cai, Chengwei Hu, and Zheng Yuan.
\newblock Cycle contrastive adversarial learning with structural consistency
  for unsupervised high-quality image deraining transformer.
\newblock {\em Neural Networks}, 178:106428, 2024.

\bibitem{he2023scalecrafter}
Yingqing He, Shaoshu Yang, Haoxin Chen, Xiaodong Cun, Menghan Xia, Yong Zhang,
  Xintao Wang, Ran He, Qifeng Chen, and Ying Shan.
\newblock Scalecrafter: Tuning-free higher-resolution visual generation with
  diffusion models.
\newblock In {\em The Twelfth International Conference on Learning
  Representations}, 2023.

\bibitem{huang2024fouriscale}
Linjiang Huang, Rongyao Fang, Aiping Zhang, Guanglu Song, Si~Liu, Yu~Liu, and
  Hongsheng Li.
\newblock Fouriscale: A frequency perspective on training-free high-resolution
  image synthesis.
\newblock In {\em European Conference on Computer Vision}, pages 196--212.
  Springer, 2024.

\bibitem{jin2023training}
Zhiyu Jin, Xuli Shen, Bin Li, and Xiangyang Xue.
\newblock Training-free diffusion model adaptation for variable-sized
  text-to-image synthesis.
\newblock {\em Advances in Neural Information Processing Systems},
  36:70847--70860, 2023.

\bibitem{kim2025diffusehigh}
Younghyun Kim, Geunmin Hwang, Junyu Zhang, and Eunbyung Park.
\newblock Diffusehigh: Training-free progressive high-resolution image
  synthesis through structure guidance.
\newblock In {\em Proceedings of the AAAI conference on artificial
  intelligence}, volume~39, pages 4338--4346, 2025.

\bibitem{zhang2024frecas}
Zhengqiang Zhang, Ruihuang Li, and Lei Zhang.
\newblock Frecas: Efficient higher-resolution image generation via
  frequency-aware cascaded sampling.
\newblock {\em The Thirteenth International Conference on Learning
  Representations}, 2025.

\bibitem{zhou2025dragflow}
Zihan Zhou, Shilin Lu, Shuli Leng, Shaocong Zhang, Zhuming Lian, Xinlei Yu, and
  Adams Wai-Kin Kong.
\newblock Dragflow: Unleashing dit priors with region based supervision for
  drag editing.
\newblock {\em arXiv preprint arXiv:2510.02253}, 2025.

\bibitem{chen2024region}
Zhennan Chen, Yajie Li, Haofan Wang, Zhibo Chen, Zhengkai Jiang, Jun Li, Qian
  Wang, Jian Yang, and Ying Tai.
\newblock Region-aware text-to-image generation via hard binding and soft
  refinement.
\newblock {\em arXiv preprint arXiv:2411.06558}, 2024.

\bibitem{du2025textcrafter}
Nikai Du, Zhennan Chen, Shan Gao, Zhizhou Chen, Xi~Chen, Zhengkai Jiang, Jian
  Yang, and Ying Tai.
\newblock Textcrafter: Accurately rendering multiple texts in complex visual
  scenes.
\newblock {\em arXiv preprint arXiv:2503.23461}, 2025.

\bibitem{gemini}
{Google}.
\newblock {Gemini}.
\newblock \url{https://gemini.google.com/}, 2025.

\bibitem{zhou2024migc}
Dewei Zhou, You Li, Fan Ma, Xiaoting Zhang, and Yi~Yang.
\newblock Migc: Multi-instance generation controller for text-to-image
  synthesis.
\newblock In {\em Proceedings of the IEEE/CVF Conference on Computer Vision and
  Pattern Recognition}, pages 6818--6828, 2024.

\bibitem{zhangrethinking}
Li~Zhang, Yan Zhong, Jianan Wang, Zhe Min, Liu Liu, et~al.
\newblock Rethinking 3d convolution in $\ell_p$-norm space.
\newblock In {\em The Thirty-eighth Annual Conference on Neural Information
  Processing Systems}, 2024.

\bibitem{zhou2024migc++}
Dewei Zhou, You Li, Fan Ma, Zongxin Yang, and Yi~Yang.
\newblock Migc++: Advanced multi-instance generation controller for image
  synthesis.
\newblock {\em IEEE Transactions on Pattern Analysis and Machine Intelligence},
  2024.

\bibitem{zhang2024vocapter}
Li~Zhang, Zean Han, Yan Zhong, Qiaojun Yu, Xingyu Wu, et~al.
\newblock Vocapter: Voting-based pose tracking for category-level articulated
  object via inter-frame priors.
\newblock In {\em ACM Multimedia 2024}, 2024.

\bibitem{lu2023tf}
Shilin Lu, Yanzhu Liu, and Adams Wai-Kin Kong.
\newblock Tf-icon: Diffusion-based training-free cross-domain image
  composition.
\newblock In {\em Proceedings of the IEEE/CVF International Conference on
  Computer Vision}, pages 2294--2305, 2023.

\bibitem{du2025upgen}
Ji~Du, Jiesheng Wu, Desheng Kong, Weiyun Liang, Fangwei Hao, Jing Xu, Bin Wang,
  Guiling Wang, and Ping Li.
\newblock Upgen: Unleashing potential of foundation models for training-free
  camouflage detection via generative models.
\newblock {\em IEEE Transactions on Image Processing}, 2025.

\bibitem{nan2024openvid}
Kepan Nan, Rui Xie, Penghao Zhou, Tiehan Fan, Zhenheng Yang, Zhijie Chen, Xiang
  Li, Jian Yang, and Ying Tai.
\newblock Openvid-1m: A large-scale high-quality dataset for text-to-video
  generation.
\newblock {\em arXiv preprint arXiv:2407.02371}, 2024.

\bibitem{si2024freeu}
Chenyang Si, Ziqi Huang, Yuming Jiang, and Ziwei Liu.
\newblock Freeu: Free lunch in diffusion u-net.
\newblock In {\em Proceedings of the IEEE/CVF Conference on Computer Vision and
  Pattern Recognition}, pages 4733--4743, 2024.

\bibitem{li2024playground}
Daiqing Li, Aleks Kamko, Ehsan Akhgari, Ali Sabet, Linmiao Xu, and Suhail
  Doshi.
\newblock Playground v2. 5: Three insights towards enhancing aesthetic quality
  in text-to-image generation.
\newblock {\em arXiv preprint arXiv:2402.17245}, 2024.

\bibitem{lu2024mace}
Shilin Lu, Zilan Wang, Leyang Li, Yanzhu Liu, and Adams Wai-Kin Kong.
\newblock Mace: Mass concept erasure in diffusion models.
\newblock In {\em Proceedings of the IEEE/CVF Conference on Computer Vision and
  Pattern Recognition}, pages 6430--6440, 2024.

\bibitem{zhou20243dis}
Dewei Zhou, Ji~Xie, Zongxin Yang, and Yi~Yang.
\newblock 3dis: Depth-driven decoupled instance synthesis for text-to-image
  generation.
\newblock {\em arXiv preprint arXiv:2410.12669}, 2024.

\bibitem{zhao2024wavelet}
Chen Zhao, Weiling Cai, Chenyu Dong, and Chengwei Hu.
\newblock Wavelet-based fourier information interaction with frequency
  diffusion adjustment for underwater image restoration.
\newblock In {\em Proceedings of the IEEE/CVF Conference on Computer Vision and
  Pattern Recognition}, pages 8281--8291, 2024.

\bibitem{zhao2024learning}
Chen Zhao, Chenyu Dong, and Weiling Cai.
\newblock Learning a physical-aware diffusion model based on transformer for
  underwater image enhancement.
\newblock {\em arXiv preprint arXiv:2403.01497}, 2024.

\bibitem{zhao2024toward}
Chen Zhao, Weiling Cai, Chenyu Dong, and Ziqi Zeng.
\newblock Toward sufficient spatial-frequency interaction for gradient-aware
  underwater image enhancement.
\newblock In {\em ICASSP 2024-2024 IEEE International Conference on Acoustics,
  Speech and Signal Processing (ICASSP)}, pages 3220--3224. IEEE, 2024.

\bibitem{chen2023diffusion}
Zhennan Chen, Rongrong Gao, Tian-Zhu Xiang, and Fan Lin.
\newblock Diffusion model for camouflaged object detection.
\newblock {\em arXiv preprint arXiv:2308.00303}, 2023.

\bibitem{DBLP:journals/corr/abs-2501-05131}
Dewei Zhou, Ji~Xie, Zongxin Yang, and Yi~Yang.
\newblock 3dis-flux: simple and efficient multi-instance generation with dit
  rendering.
\newblock {\em CoRR}, abs/2501.05131, 2025.

\bibitem{DBLP:journals/corr/abs-2504-15376}
Zhiqiu Lin, Siyuan Cen, Daniel Jiang, Jay Karhade, Hewei Wang, Chancharik
  Mitra, Tiffany Ling, Yuhan Huang, Sifan Liu, Mingyu Chen, Rushikesh Zawar,
  Xue Bai, Yilun Du, Chuang Gan, and Deva Ramanan.
\newblock Towards understanding camera motions in any video.
\newblock {\em CoRR}, abs/2504.15376, 2025.

\bibitem{xie2025star}
Rui Xie, Yinhong Liu, Penghao Zhou, Chen Zhao, Jun Zhou, Kai Zhang, Zhenyu
  Zhang, Jian Yang, Zhenheng Yang, and Ying Tai.
\newblock Star: Spatial-temporal augmentation with text-to-video models for
  real-world video super-resolution.
\newblock {\em arXiv preprint arXiv:2501.02976}, 2025.

\bibitem{du2025shift}
Ji~Du, Fangwei Hao, Mingyang Yu, Desheng Kong, Jiesheng Wu, Bin Wang, Jing Xu,
  and Ping Li.
\newblock Shift the lens: Environment-aware unsupervised camouflaged object
  detection.
\newblock In {\em Proceedings of the Computer Vision and Pattern Recognition
  Conference}, pages 19271--19282, 2025.

\bibitem{zhao2025multi}
Chen Zhao, Wei-Ling Cai, Zheng Yuan, and Cheng-Wei Hu.
\newblock Multi-cropping contrastive learning and domain consistency for
  unsupervised image-to-image translation.
\newblock {\em IET Image Processing}, 19(1):e70006, 2025.

\bibitem{ho2020denoising}
Jonathan Ho, Ajay Jain, and Pieter Abbeel.
\newblock Denoising diffusion probabilistic models.
\newblock {\em Advances in neural information processing systems},
  33:6840--6851, 2020.

\bibitem{song2020denoising}
Jiaming Song, Chenlin Meng, and Stefano Ermon.
\newblock Denoising diffusion implicit models.
\newblock {\em arXiv preprint arXiv:2010.02502}, 2020.

\bibitem{rombach2022high}
Robin Rombach, Andreas Blattmann, Dominik Lorenz, Patrick Esser, and Bj{\"o}rn
  Ommer.
\newblock High-resolution image synthesis with latent diffusion models.
\newblock In {\em Proceedings of the IEEE/CVF conference on computer vision and
  pattern recognition}, pages 10684--10695, 2022.

\bibitem{zhuo2024lumina}
Le~Zhuo, Ruoyi Du, Han Xiao, Yangguang Li, Dongyang Liu, Rongjie Huang, Wenze
  Liu, Lirui Zhao, Fu-Yun Wang, Zhanyu Ma, et~al.
\newblock Lumina-next: Making lumina-t2x stronger and faster with next-dit.
\newblock {\em arXiv preprint arXiv:2406.18583}, 2024.

\bibitem{zhou2023pyramid}
Dewei Zhou, Zongxin Yang, and Yi~Yang.
\newblock Pyramid diffusion models for low-light image enhancement.
\newblock {\em arXiv preprint arXiv:2305.10028}, 2023.

\bibitem{chai2022any}
Lucy Chai, Michael Gharbi, Eli Shechtman, Phillip Isola, and Richard Zhang.
\newblock Any-resolution training for high-resolution image synthesis.
\newblock In {\em European conference on computer vision}, pages 170--188.
  Springer, 2022.

\bibitem{zhang2025u}
Li~Zhang, Weiqing Meng, Yan Zhong, Bin Kong, Mingliang Xu, Jianming Du, Xue
  Wang, Rujing Wang, and Liu Liu.
\newblock U-cope: Taking a further step to universal 9d category-level object
  pose estimation.
\newblock In {\em European Conference on Computer Vision}, pages 254--270.
  Springer, 2025.

\bibitem{zhou2025dreamrenderer}
Dewei Zhou, Mingwei Li, Zongxin Yang, and Yi~Yang.
\newblock Dreamrenderer: Taming multi-instance attribute control in large-scale
  text-to-image models.
\newblock {\em arXiv preprint arXiv:2503.12885}, 2025.

\bibitem{lu2024robust}
Shilin Lu, Zihan Zhou, Jiayou Lu, Yuanzhi Zhu, and Adams Wai-Kin Kong.
\newblock Robust watermarking using generative priors against image editing:
  From benchmarking to advances.
\newblock {\em arXiv preprint arXiv:2410.18775}, 2024.

\bibitem{zhang2024catmullrom}
Li~Zhang, Mingliang Xu, Dong Li, Jianming Du, and Rujing Wang.
\newblock Catmullrom splines-based regression for image forgery localization.
\newblock In {\em Proceedings of the AAAI Conference on Artificial
  Intelligence}, volume~38, pages 7196--7204, 2024.

\bibitem{lu2025does}
Shilin Lu, Zhuming Lian, Zihan Zhou, Shaocong Zhang, Chen Zhao, and Adams
  Wai-Kin Kong.
\newblock Does flux already know how to perform physically plausible image
  composition?
\newblock {\em arXiv preprint arXiv:2509.21278}, 2025.

\bibitem{schuhmann2022laion}
Christoph Schuhmann, Romain Beaumont, Richard Vencu, Cade Gordon, Ross
  Wightman, Mehdi Cherti, Theo Coombes, Aarush Katta, Clayton Mullis, Mitchell
  Wortsman, et~al.
\newblock Laion-5b: An open large-scale dataset for training next generation
  image-text models.
\newblock {\em Advances in neural information processing systems},
  35:25278--25294, 2022.

\bibitem{yitowards}
Mingyang Yi, Aoxue Li, Yi~Xin, and Zhenguo Li.
\newblock Towards understanding the working mechanism of text-to-image
  diffusion model.
\newblock In {\em The Thirty-eighth Annual Conference on Neural Information
  Processing Systems}.

\bibitem{zhang2021designing}
Kai Zhang, Jingyun Liang, Luc Van~Gool, and Radu Timofte.
\newblock Designing a practical degradation model for deep blind image
  super-resolution.
\newblock In {\em Proceedings of the IEEE/CVF international conference on
  computer vision}, pages 4791--4800, 2021.

\bibitem{du2024max}
Ruoyi Du, Dongyang Liu, Le~Zhuo, Qin Qi, Hongsheng Li, Zhanyu Ma, and Peng Gao.
\newblock I-max: Maximize the resolution potential of pre-trained rectified
  flow transformers with projected flow.
\newblock {\em arXiv preprint arXiv:2410.07536}, 2024.

\bibitem{zhang2024long}
Beichen Zhang, Pan Zhang, Xiaoyi Dong, Yuhang Zang, and Jiaqi Wang.
\newblock Long-clip: Unlocking the long-text capability of clip.
\newblock In {\em European Conference on Computer Vision}, pages 310--325.
  Springer, 2024.

\bibitem{xie2025fg}
Chunyu Xie, Bin Wang, Fanjing Kong, Jincheng Li, Dawei Liang, Gengshen Zhang,
  Dawei Leng, and Yuhui Yin.
\newblock Fg-clip: Fine-grained visual and textual alignment.
\newblock {\em arXiv preprint arXiv:2505.05071}, 2025.

\bibitem{heusel2017gans}
Martin Heusel, Hubert Ramsauer, Thomas Unterthiner, Bernhard Nessler, and Sepp
  Hochreiter.
\newblock Gans trained by a two time-scale update rule converge to a local nash
  equilibrium.
\newblock {\em Advances in neural information processing systems}, 30, 2017.

\bibitem{salimans2016improved}
Tim Salimans, Ian Goodfellow, Wojciech Zaremba, Vicki Cheung, Alec Radford, and
  Xi~Chen.
\newblock Improved techniques for training gans.
\newblock {\em Advances in neural information processing systems}, 29, 2016.

\end{thebibliography}
\bibliographystyle{unsrt}

\end{document}